\documentclass[journal]{IEEEtran}

\ifCLASSINFOpdf
\else
\fi

\usepackage{amsmath}
\usepackage{amssymb}
\usepackage{dblfloatfix}
\usepackage{booktabs}
\usepackage{enumitem}
\usepackage{multirow}
\usepackage{tikz}
\usepackage{url}
\usepackage{subcaption}
\usepackage{color}
\usepackage{adjustbox}
\usepackage{caption} %
\usepackage[english]{datetime2}
\usepackage{textcomp}
\usepackage[T1]{fontenc}
\usepackage{tabulary}
\usepackage{graphicx}
\usepackage{refcount}
\usepackage{nicefrac}
\usepackage{tablefootnote}

\newcommand{\etal}{\textit{et al}.\@ }
\newcommand{\ie}{\textit{i}.\textit{e}.\@ }
\newcommand{\eg}{\textit{e}.\textit{g}.\@ }
\newcommand{\etc}{\textit{etc}.\@}

\newcommand{\norm}[1]{\left\lVert#1\right\rVert}

\begin{document}
\title{A Review of Automated Speech and Language Features for Assessment of Cognitive and Thought Disorders}

\author{Rohit~Voleti,~\IEEEmembership{Student~Member,~IEEE,} Julie~M.~Liss,
        Visar~Berisha,~\IEEEmembership{Member,~IEEE,}%
\thanks{R. Voleti is with the School
of Electrical, Computer, \& Energy Engineering, Arizona State University, Tempe,
AZ, 85281 USA e-mail: rnvoleti@asu.edu}%
\thanks{Manuscript received 16 May 2019; revised 23 Aug 2019; revised 24 Oct 2019; Accepted 26 Oct 2019}}

\markboth{IEEE Journal of Selected Topics in Signal Processing (JSTSP), Special Issue}%
{Shell \MakeLowercase{\textit{et al.}}: Bare Demo of IEEEtran.cls for IEEE Journals}

\IEEEpubid{J-STSP-AAHD-00183-2019~\copyright~2019 IEEE}

\maketitle

\begin{abstract}
It is widely accepted that information derived from analyzing speech (the acoustic signal) and language production (words and sentences) serves as a useful window into the health of an individual's cognitive ability.
In fact, most neuropsychological testing batteries have a component related to speech and language where clinicians elicit speech from patients for subjective evaluation across a broad set of dimensions.
With advances in speech signal processing and natural language processing, there has been recent interest in developing tools to detect more subtle changes in cognitive-linguistic function. 
This work relies on extracting a set of features from recorded and transcribed speech for objective assessments of speech and language, early diagnosis of neurological disease, and tracking of disease after diagnosis.
With an emphasis on cognitive and thought disorders, in this paper we provide a review of existing speech and language features used in this domain, discuss their clinical application, and highlight their advantages and disadvantages.
Broadly speaking, the review is split into two categories: language features based on natural language processing and speech features based on speech signal processing.
Within each category, we consider features that aim to measure complementary dimensions of cognitive-linguistics, including language diversity, syntactic complexity, semantic coherence, and timing.
We conclude the review with a proposal of new research directions to further advance the field.   
\end{abstract}


\section{Introduction}

\IEEEPARstart{E}{arly} detection of neurodegenerative disease and episodes of mental illness that impact cognitive function (henceforth, \emph{cognitive disorder} and \emph{thought disorder}, respectively) is a major goal of current research trends in speech and language processing.
These afflictions have both significant societal and economic impacts on affected individuals.
It is estimated that approximately one in six adults in the United States lives with some form of mental illness, according to the National Institute of Mental Health (NIMH), totaling 44.6 million people in 2016 \cite{center20172016}. 
In the United States alone, some estimate that the economic burden of mental illness is approximately \$1 trillion annually~\cite{cecchiComputingStructureLanguage2017}.
The World Health Organization estimates that, in 2015, the global burden of Alzheimer's disease and dementia equaled $1.1$\% of the global gross domestic product~\cite{princeGlobalImpactDementia2015}.
The ability to identify early signs and symptoms is critical for the development of interventions to impact progression or episodes. 
\begin{figure}
	\centering
	\includegraphics[width=0.80\linewidth]{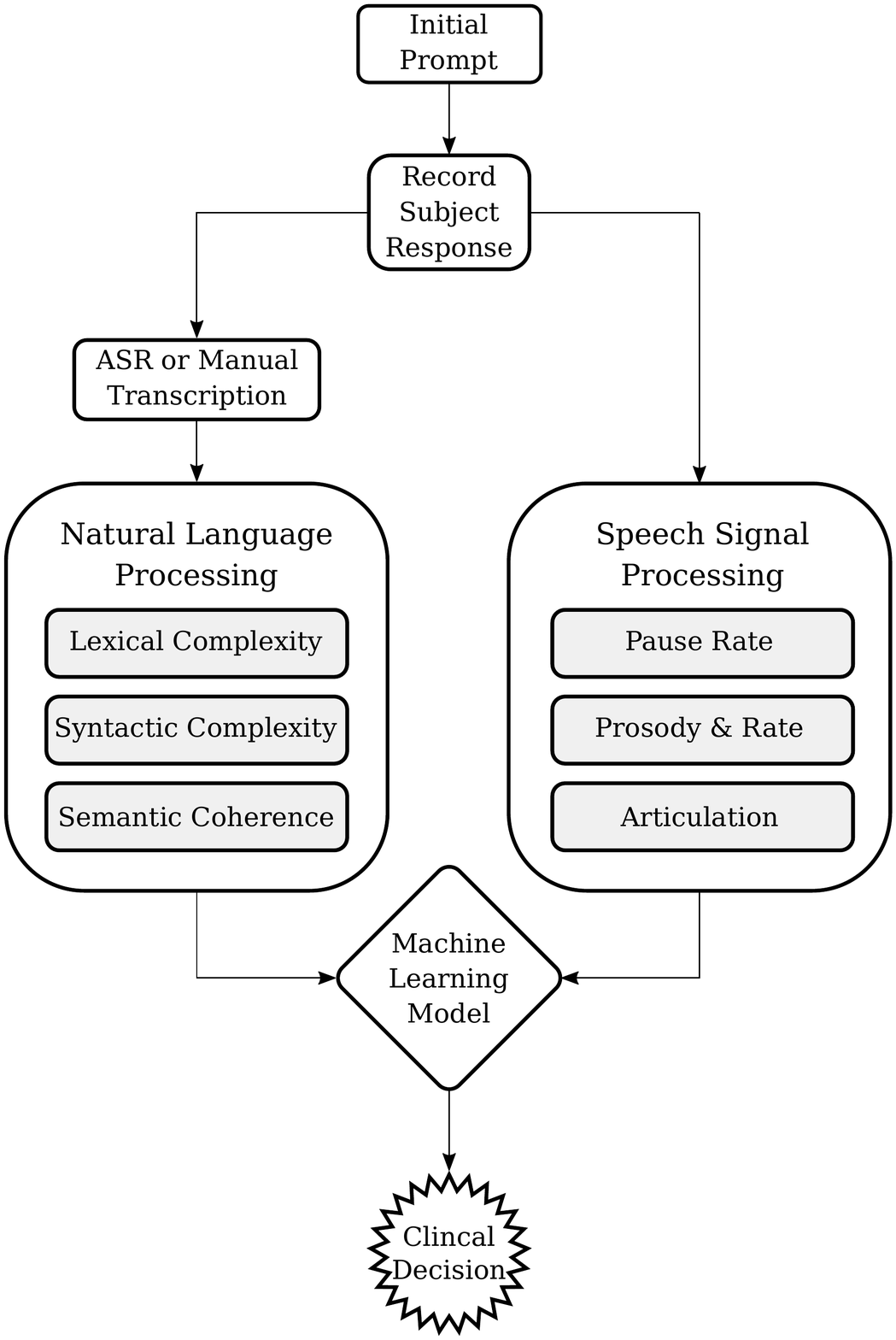}
	\caption{Overview of the process of using natural language processing and speech signal processing for extraction of speech and language features for clinical decision-making. Example language features include lexical complexity, syntactic complexity, semantic coherence, \etc
		~Example of acoustic speech features include pause rate, prosody, articulation, \etc}
	\label{fig:overview}
\end{figure}

\begin{table*}[]
\resizebox{\textwidth}{!}{%
	\begin{tabular}{llll}
		\hline
		Category   & Subcategory          & Features or Methods Used                                                                                                                            & Cognitive \& Thought Disorder(s) Assessed                                                                                                                                                                                                                                                       \\ \hline
		Text-based & Lexical features     & Bag of words vocabulary analysis                                                                                                                    & Semantic dementia (SD)~\cite{garrardMachineLearningApproaches2014}; Alzheimer's disease (AD)~\cite{fraserLinguisticFeaturesIdentify2015}                                                                                                                                                        \\
		&                      & Linguistic Inquiry \& Word Count (LIWC)~\cite{tausczik2010psychological}                                                                            & Mild cognitive impairment (MCI)~\cite{asgariPredictingMildCognitive2017}, schizophrenia (Sz/Sza)~\cite{mitchell2015quantifying}; AD~\cite{sadeghianSpeechProcessingApproach2017}                                                                                                                \\
		&                      & Lexical Diversity  (TTR, MATTR, BI, HS, \etc)                                                                                                       & AD~\cite{fraserLinguisticFeaturesIdentify2015, bucksAnalysisSpontaneousConversational2000, berishaTrackingDiscourseComplexity2015}; primary progressive aphasia (PPA)~\cite{fraserAutomatedClassificationPrimary2012},                                                                          \\
		&                      &                                                                                                                                                     & Sz/Sza~\cite{Voleti2019}, bipolar disorder (BPD)~\cite{Voleti2019}                                                                                                                                                                          \\
		&                      & Lexical Density  (content density, idea density, $P$-density)                                                                                       & MCI~\cite{roarkSpokenLanguageDerived2011}; AD~\cite{fraserLinguisticFeaturesIdentify2015, berishaTrackingDiscourseComplexity2015, sadeghianSpeechProcessingApproach2017}; PPA~\cite{fraserAutomatedClassificationPrimary2012};                                                                  \\
		&                      &                                                                                                                                                     & chronic traumatic encephalopathy (CTE)~\cite{berishaLongitudinalChangesLinguistic2017}; Sz/Sza~\cite{Voleti2019}, BPD~\cite{Voleti2019}                                                                                                     \\
		&                      & Part-of-speech (POS) tagging                                                                                                                        & MCI~\cite{roarkSpokenLanguageDerived2011}; PPA~\cite{fraserAutomatedClassificationPrimary2012}; AD~\cite{fraserLinguisticFeaturesIdentify2015, sadeghianSpeechProcessingApproach2017}; Sz/Sza~\cite{bediAutomatedAnalysisFree2015, corcoranPredictionPsychosisProtocols2018, iter2018automatic} \\
		&                      &                                                                                                                                                     &                                                                                                                                                                                                                                                                                                 \\
		& Syntactical features & Constituency-based parse tree scores (Yngve~\cite{yngveModelHypothesisLanguage1960}, Frazier~\cite{frazierSyntacticComplexity1985})                 & AD~\cite{fraserLinguisticFeaturesIdentify2015, sadeghianSpeechProcessingApproach2017}; MCI~\cite{roarkSpokenLanguageDerived2011}; PPA~\cite{fraserAutomatedClassificationPrimary2012}                                                                                                           \\
		&                      & Dependency-based parse tree scores                                                                                                                  & MCI~\cite{roarkSpokenLanguageDerived2011}                                                                                                                                                                                                                                                       \\
		&                      & Speech graphs and attributes                                                                                                                        & Sz/Sza~\cite{motaSpeechGraphsProvide2012, motaGraphAnalysisDream2014}; BPD~\cite{motaSpeechGraphsProvide2012, motaGraphAnalysisDream2014}; MCI~\cite{bertolaGraphAnalysisVerbal2014}                                                                                                            \\
		&                      &                                                                                                                                                     &                                                                                                                                                                                                                                                                                                 \\
		& Semantic features    & Word \& sentence embeddings:                                                                                                                        &                                                                                                                                                                                                                                                                                                 \\
		&                      & ~- LSA~\cite{landauerSolutionPlatoProblem1997}                                                                                                      & Sz/Sza~\cite{elvevagQuantifyingIncoherenceSpeech2007, bediAutomatedAnalysisFree2015, corcoranPredictionPsychosisProtocols2018}                                                                                                                                                                  \\
		&                      & ~- Neural word embeddings (\emph{word2vec}~\cite{mikolovEfficientEstimationWord2013},  \emph{GloVe}~\cite{penningtonGloveGlobalVectors2014}, \etc)  & Sz/Sza~\cite{kayiPredictiveLinguisticFeatures2017, iter2018automatic, Voleti2019}; BPD~\cite{Voleti2019}                                                                                                                                    \\
		&                      & ~- Neural sentence embeddings (SIF~\cite{aroraSimpleToughtoBeatBaseline2017}, \emph{InferSent}~\cite{conneauSupervisedLearningUniversal2017}, \etc) & Sz/Sza~\cite{iter2018automatic, Voleti2019}; BPD~\cite{Voleti2019}                                                                                                                                                                          \\
		&                      & Topic modeling:                                                                                                                                     &                                                                                                                                                                                                                                                                                                 \\
		&                      & ~- LDA~\cite{bleiLatentDirichletAllocation2003}                                                                                                     & Sz/Sza~\cite{kayiPredictiveLinguisticFeatures2017, mitchell2015quantifying}                                                                                                                                                                                                                     \\
		&                      & ~- Vector-space topic modeling with neural networks                                                                                                 & AD~\cite{yanchevaVectorspaceTopicModels2016, hernandez-dominguezComputerbasedEvaluationAlzheimer2018}; MCI~\cite{hernandez-dominguezComputerbasedEvaluationAlzheimer2018}                                                                                                                       \\
		&                      & Semantic role labeling~\cite{dasProbabilisticFramesemanticParsing2010}                                                                              & Sz/Sza~\cite{kayiPredictiveLinguisticFeatures2017}                                                                                                                                                                                                                                              \\
		&                      &                                                                                                                                                     &                                                                                                                                                                                                                                                                                                 \\
		& Pragmatics           & Sentiment analysis                                                                                                                                  & Sz/Sza~\cite{kayiPredictiveLinguisticFeatures2017}                                                                                                                                                                                                                                              \\
		&                      &                                                                                                                                                     &                                                                                                                                                                                                                                                                                                 \\
		Acoustic   & Prosodic features    & Temporal (pause rate, phonation rate, voiced durations, \etc)                                                                                       & MCI~\cite{roarkSpokenLanguageDerived2011, konigAutomaticSpeechAnalysis2015, tothAutomaticDetectionMild2015, tothSpeechRecognitionbasedSolution2018}; AD~\cite{konigAutomaticSpeechAnalysis2015, lopez-de-ipinaAutomaticDiagnosisAlzheimer2015}; Sz/Sza~\cite{tahirNonverbalSpeechAnalysis2016}  \\
		&                      &                                                                                                                                                     & Frontotemporal lobal degeneration (FTLD)~\cite{pakhomovComputerizedAnalysisSpeech2010}                                                                                                                                                                                                          \\
		&                      & Fundamental frequency ($F_0$) and trajectory                                                                                                        & AD~\cite{lopez-de-ipinaAutomaticDiagnosisAlzheimer2015}; BPD~\cite{guidiAutomaticAnalysisSpeech2015}                                                                                                                                                                                            \\
		&                      & Loudness and energy                                                                                                                                 & AD~\cite{lopez-de-ipinaAutomaticDiagnosisAlzheimer2015}                                                                                                                                                                                                                                         \\
		&                      & Emotional content                                                                                                                                   & AD~\cite{lopez-de-ipinaAutomaticDiagnosisAlzheimer2015}                                                                                                                                                                                                                                         \\
		&                      &                                                                                                                                                     &                                                                                                                                                                                                                                                                                                 \\
		& Spectral features    & Formant trajectories ($F_1$, $F_2$, $F_3$, \etc)                                                                                                    & PPA~\cite{fraserUsingTextAcoustic2013}; AD~\cite{fraserLinguisticFeaturesIdentify2015}                                                                                                                                                                                                          \\
		&                      & Spectral centroid~\cite{peeters2004large}                                                                                                           & AD~\cite{lopez-de-ipinaAutomaticDiagnosisAlzheimer2015}                                                                                                                                                                                                                                         \\
		&                      & MFCC statistics~\cite{davisComparisonParametricRepresentations1980}                                                                                 & PPA~\cite{fraserUsingTextAcoustic2013}; AD~\cite{fraserLinguisticFeaturesIdentify2015}                                                                                                                                                                                                          \\
		&                      &                                                                                                                                                     &                                                                                                                                                                                                                                                                                                 \\
		& Vocal quality        & Jitter, shimmer, harmonic-to-noise ratio (HNR)                                                                                                      & PPA~\cite{fraserUsingTextAcoustic2013}; AD~\cite{fraserLinguisticFeaturesIdentify2015, lopez-de-ipinaAutomaticDiagnosisAlzheimer2015}                                                                                                                                                           \\
		&                      &                                                                                                                                                     &                                                                                                                                                                                                                                                                                                 \\
		& ASR-related          & Phone-level detection of filled pauses \& temporal features                                                                                         & MCI~\cite{tothAutomaticDetectionMild2015, tothSpeechRecognitionbasedSolution2018}                                                                                                                                                                                                               \\
		&                      & Improving $\mathrm{WER}$ for clinical data                                                                                                          & AD~\cite{zhouSpeechRecognitionAlzheimer2016, sadeghianSpeechProcessingApproach2017}; neurodegenerative dementia (ND)~\cite{mirheidariDiagnosingPeopleDementia2016, mirheidariAutomationDiagnosticConversation2017, Weiner2017}                                                                  \\ \hline
	\end{tabular}%
}
\caption{Summary of work covered in this review, including textual and acoustic features to assess cognitive and thought disorders that are automatically extracted from speech and language samples using NLP and speech signal processing. Note that in the abbreviations in the table, ``Sz'' refers to schizophrenia and ``Sza'' refers to the related schizoaffective disorder, which are considered together for simplicity in this summary.}
\label{tab:summary}
\end{table*}

\IEEEpubidadjcol
Many aspects of cognitive and thought disorders are manifest in the way speech is produced and what is said.
Irrespective of the underlying disease or condition, the analysis of speech and language can provide insight to the underlying neural function.
This has motivated current research trends in quantitative speech and language analytics, with the hope of eventually developing clinically-validated algorithms for better diagnosis, prediction, and characterization of these conditions.
This has both long-term and nearer-term potential for impact.
In the long term, there is the potential for new diagnostics and early intervention for improved treatment outcomes and reduced  economic burden.
In the nearer term, there is the potential for improving the efficiency of clinical trials evaluating new drugs.
It is generally accepted that early enrollment in clinical trials for evaluation of new drugs maximizes the chances of showing that a drug is successful~\cite{jerominBiomarkersNeurodegenerativeDiseases2017, katsunoPreclinicalProgressionNeurodegenerative2018}.
In addition, adopting endpoints that are more sensitive to change means that these studies can be powered with fewer participants.
Digital endpoints collected frequently have recently garnered interest in this domain~\cite{dodgeUseHighFrequencyInHome2015}.

In this review, we limit our focus to speech and language analysis for \emph{cognitive} and \emph{thought disorders} in the context of neurodegenerative disease and mental illness.
In keeping with the \emph{Diagnostic and Statistical Manual of Mental Disorders} (DSM-5)~\cite{americanpsychiatricassociationDiagnosticStatisticalManual2013} classification system, ``cognitive disorders'' refer to disturbances in memory and cognition, whereas ``thought disorders'' refer to the inability to produce and sustain coherent communication. 
Thus, Alzheimer's disease (the most common form of dementia) is an example of a cognitive disorder, while schizophrenia is an example of a mental illness that presents as a thought disorder.

With access to clinical speech and language databases along with recent developments in the fields of speech signal processing, computational linguistics, and machine learning, there is an increased potential for using computational methods to automate the analysis of speech and language datasets for clinical applications~\cite{cecchiComputingStructureLanguage2017}.
Objective analysis of this sort has the potential to overcome some of the limitations associated with the current state-of-the-art for improved diagnosis, prediction, and characterization of cognitive and thought disorders. 
A high-level block diagram of existing methods in clinical-speech analytics is shown in Figure~\ref{fig:overview}. 
Patients provide speech samples via a speech elicitation task.
This could be passively collected speech, patient interviews, or recorded neuropsychological batteries.
The resulting speech is transcribed, using either automatic speech recognition (ASR) or manual transcription, and a set of speech and language features are extracted that aim to measure different aspects of cognitive-linguistic change.
These features become the input of a machine learning model that aims to predict a dependent variable of interest, \eg detection of clinical conditions or assessment of social skills~\cite{pattersonSocialSkillsPerformance2001, roarkSpokenLanguageDerived2011, Voleti2019}.

Perhaps the most important parts of the analysis framework in Figure~\ref{fig:overview} are the analytical methods used to extract clinically-relevant features from the samples. 
With a focus on cognitive and thought disorders, we provide a survey of the existing literature and common speech and language features used in this context. 
The review is not focused on a particular disease, but rather on the methods for extracting clinically-relevant measures from the resultant speech. 

A summary of the work reviewed in this paper can be seen in Table~\ref{tab:summary} and will be discussed in the subsequent sections.
In the next section, we aim to place our review in context. We provide an overview of speech and language production, highlight the existing neuropsychological assessments used to evaluate speech and language, and provide an overview of the the speech and language dimensions that we focus on in this review.

Following that section, the review is split into two parts: natural language processing (NLP) features and speech signal processing features.
With NLP, we can measure the complexity and coherence of language and with speech signal processing, we can measure acoustic proxies related to cognitive processing. 

Finally, in Section~\ref{sec:discussion} we discuss gaps in current research, propose future directions, and end with concluding remarks.

\section{Background}

\subsection{Spoken Language Production}
\begin{figure*}
	\centering
	\includegraphics[width=0.83\linewidth]{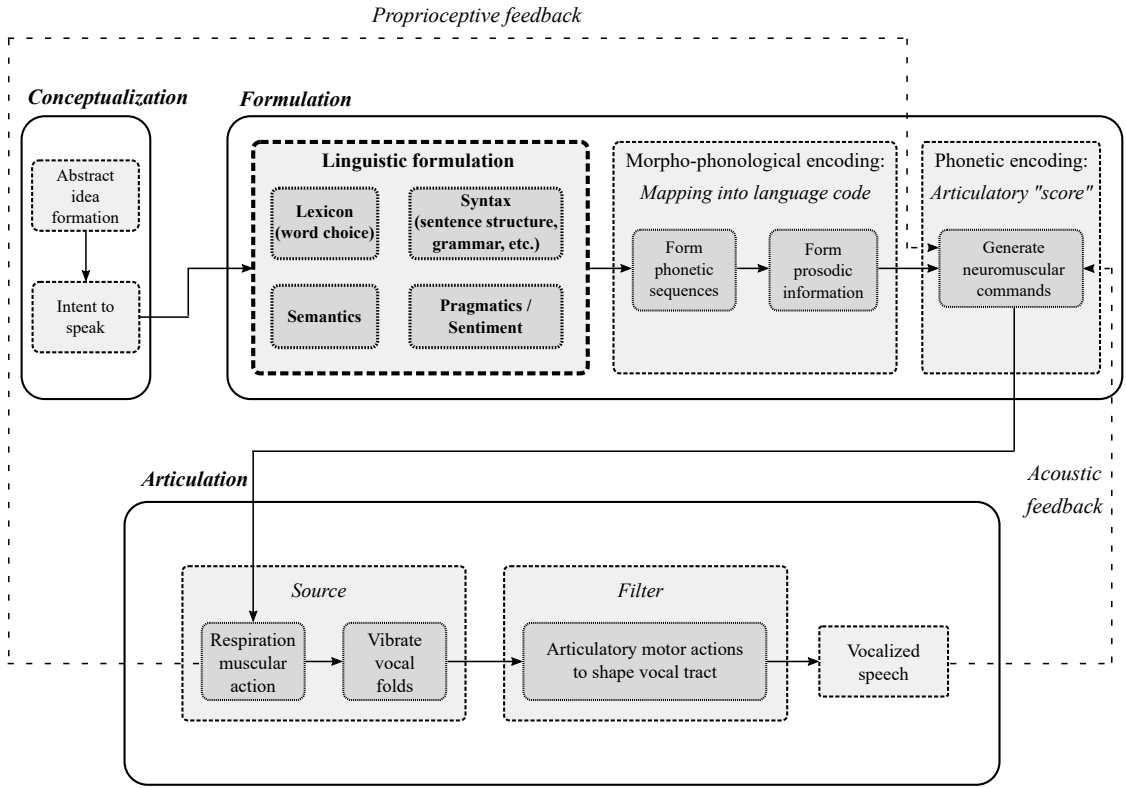}
	\caption{Speech production block diagram model, adapted and modified from~\protect\cite{cumminsReviewDepressionSuicide2015}. In this review, we focus primarily on the additional box we termed ``Linguistic Formulation'' within the formulation stage of speech production. Cognitive and thought disorders that affect this area have direct measurable outputs on the actual language content which can be studied by statistical text-based analysis. Additionally, they also have indirect downstream effects on the vocalized and articulated speech acoustics. Both of these areas are covered in our review.}
	\label{fig:production}
\end{figure*}

The production of spoken language in humans is a complex, multi-stage process that involves high levels of memory, cognition, and sensorimotor function.
There are three distinct stages~\cite{Levelt1999}:
\begin{enumerate}
	\item \emph{Conceptualization}: involves the formation of abstract ideas about the intended message to be communicated
	\item \emph{Formulation}: involves forming the exact linguistic construction of the utterance to be spoken
	\item \emph{Articulation}: involves actually producing sounds using the various components of the speech production system, \ie lungs, glottis, larynx, vocal tract, \etc
\end{enumerate}
These stages are visually represented in the block diagram in Figure~\ref{fig:production}.
In the conceptualization stage, pre-verbal ideas are formed to link a desired concept to be expressed to the spoken language that is eventually formed.
The formulation stage consists of several distinct components: (a) lexical, syntactical, \& grammatical formulations, (b) morpho-phonological encoding, and (c) phonetic encoding.
This involves forming the linguistic structure of a spoken utterance, determining which syllables are needed to articulate the utterance, and the creation of an \emph{articulatory score} containing instructions that are to be executed by the vocal apparatus in the articulation stage~\cite{Levelt1999}.

Cognitive and thought disorders have the ability to affect any of these stages, but broadly, they can be captured through analysis of ``content'' (what is said) and ``form'' (how it is said). 
Indeed, the tools used to characterize content and form of speech are agnostic to the underlying condition. 
It is the constellation of features shown to be affected that converge on the locus of deficit for an individual.
For example, speech that lacks coherence of ideas and jumps from topic to topic (impaired content), and is produced very rapidly and without pauses (impaired form), would point toward a thought or mood disorder, such as schizophrenia or mania.
A person with dementia may present with reduced vocabulary size (impaired content), and with increased number and duration of pauses (impaired form).
To reiterate, the speech and language measures are, themselves, agnostic to the underlying disorder.
Rather, it is the constellation of deficit patterns that associate with different etiologies. Our aim here is to provide an overview of the methods used to extract these constellations without focusing on a particular disease area. We refer to clinical applications in each section to highlight existing work that uses these features in clinical applications. 
 
\subsection{Clinical Assessment of Speech  \& Language for Cognitive \& Thought Disorders}

A variety of clinical protocols exist for the evaluation and diagnosis of disorders affecting cognitive function in psychiatry and neurology.
The DSM-5~\cite{americanpsychiatricassociationDiagnosticStatisticalManual2013}, published by the American Psychiatric Association (APA), provides the standard diagnostic criteria for psychiatric and neurocognitive disorders, and it is updated as knowledge in the field evolves.
The DSM-5 covers a large spectrum of psychiatric and cognitive disorders, such as depression, anxiety, schizophrenia, bipolar disorder, dementia, and several more.

Based on these criteria, many evaluation methodologies have been developed in clinical practice for diagnosing and evaluating these cognitive disorders.
The \emph{mental status examination} (MSE) is a commonly utilized and multi-faceted tool for screening an individual at a given point in time for signs of neurological and psychiatric disorders~\cite{trzepaczPsychiatricMentalStatus1993}.
Components of the MSE evaluate affect, mood, appearance, judgment, speech, thought process, and overall cognitive ability through a variety of tasks and surveys.
Related screenings include the \emph{mini-mental state examination} (MMSE)~\cite{pangmanExaminationPsychometricProperties2000}, \emph{Addenbrooke's Cognitive Examination} (ACE)~\cite{mathuranathBriefCognitiveTest2000}, and \emph{the Montreal Cognitive Assessment} (MoCA)~\cite{nasreddineMontrealCognitiveAssessment2005} for evaluating conditions like mild cognitive impairment (MCI), dementia, and Alzheimer's disease (AD).
Other forms of disorders, \ie schizophrenia and bipolar disorder, can be evaluated with screenings such as the \emph{Clinical Assessment Interview for Negative Symptoms} (CAINS)~\cite{kringClinicalAssessmentInterview2013}, \emph{Brief Negative Symptom Scale} (BNSS)~\cite{kirkpatrickBriefNegativeSymptom2011}, the \emph{Social Skills Performance Assessment} (SSPA)~\cite{pattersonSocialSkillsPerformance2001}, and several others that measure the effects of thought and mood disorders.

All of these neuropsychological batteries for evaluating cognitive health have a significant speech and language assessment, as cognitive-linguistic function is a strong biomarker for neuropsychological health in many dimensions.
However, ratings for narrative, recall, conversational, or other spoken language tasks are often subjective in nature and of variable reliability, making the underlying diagnosis more challenging ~\cite{tzurbitanAttitudesMentalHealth2018}.
While the diagnosis for many common psychiatric conditions has become more consistent over time as they are better understood, others (\eg schizoaffective disorder) are often evaluated inconsistently by different clinical assessors due to the subjective nature of the test batteries applied~\cite{harveyImpairedSelfassessmentSchizophrenia2015, piesHowObjectiveAre2007}.
The speech and language samples collected during these screenings serve as potentially valuable databases for \emph{objective} and automatically computable measures of cognitive-linguistic ability.
Recent research suggests that analysis of this rich data allows us to explore several new objective dimensions for evaluation, which has a largely untapped potential to improve clinical assessment and outcomes. 
These new tools have the potential to provide a finer-grained analysis of the resultant speech when compared against existing rating scales.

\subsection{Speech \& Language Dimensions of Interest}
Natural spoken language contains several measurable dimensions that indicate various aspects of cognitive health.
In this review, we are interested in the analysis of linguistic and acoustic speech features that are indicative of cognitive and thought disorders related to \emph{thought content} and \emph{thought organization}.
These include a variety of neurological impairments (\eg MCI, dementia, AD, chronic traumatic encephalopathy) and psychiatric conditions (\eg schizophrenia, schizoaffective disorder, bipolar disorder).

Most of the work in this space exists in the context of textual language analysis, either by manual or automatic transcription of spoken utterances.
Looking at Figure~\ref{fig:production}, we focus mainly on the ``\emph{Linguistic formulation}'' area within the formulation stage.
Neurological thought disorders all affect the ability of an individual to form complex thoughts and sentence structures, and may often have issues such as \emph{poverty of speech} or disorganized speech.
Therefore, we look at methods for examining thought content density, complexity of sentence syntax, semantic coherence, and sentiment analysis as they relate to these conditions.

Analysis of acoustic speech samples leads to additional insight for characterizing  neurological and psychiatric thought disorders, as impairments in language formation in turn affect the articulation of the spoken output.
As seen in Figure~\ref{fig:production}, the articulation pathway that leads to speech output depends upon the cognitive ability required for the conceptualization and formulation stages of speech production.
Cognitive and working memory disorders can lead to impairments in neuromuscular motor coordination and proprioceptive feedback as well, affecting speech output~\cite{krajewskiApplyingMultipleClassifiers2012}. 
Among the speech signal features considered are those related to temporal analysis and prosody (\eg pause rate, phonation rate, periodicity of speech, \etc) and those related to frequency analysis (\eg mean, variance, kurtosis of Mel frequency cepstral coefficients).

We note that the purpose of this review is to highlight recent research that identifies and characterizes automatically computed speech and language features related to neurological and psychiatric disorders of thought content and formulation.
In each part, we will provide an overview of commonly used techniques for extracting various speech and language features, present examples of their clinical application, and discuss the advantages and disadvantages of the methods reviewed.

\section{Measuring Cognitive and Thought Disorders with Natural Language Processing}\label{sec:nlp}
In this section, we will provide a review of several families of natural language processing methods that range from simple lexical analysis to state-of-the-art language models that can be utilized for clinical assessment.

The sections below present families of approaches in order of increasing complexity. 
In the first section, we describe methods based on subjective evaluation of speech and language; then we discuss methods that rely on lexeme-level information, followed by methods that rely on sentence-level information, and end with methods that rely on semantics.
For each section, we provide a description of representative approaches and a review of how these methods are used in clinical applications. We end each section with a discussion of the advantages and disadvantages of the approaches in that section.

\subsection{Early Work} \label{subsec:nun}
Simple analysis of written language samples has long been thought to provide valuable information regarding cognitive health. One of the best-known early examples of such work is the famous ``nun study'' by Snowdon \etal on linguistic ability as a predictor of Alzheimer's disease (AD)~\cite{10.1001/jama.1996.03530310034029}.
In this work, manual evaluations of the linguistic abilities of 93 nuns were conducted by analysis of autobiographical essays they had written earlier in their lives.
The researchers evaluated the linguistic structure of the essays by scoring the grammatical complexity and idea density in the writing samples.
In particular, the study found that low idea density in early life was a particularly strong predictor of reduced cognitive ability or the presence of AD in later life.
Roughly $80$\% of the participants that were determined to lack linguistic complexity in their writings developed AD or had mental and cognitive disabilities in their older age.

This work was groundbreaking in showing that linguistic structure and complexity can serve as a strong predictor for the onset of AD and potentially other forms of cognitive impairment. 
However, it required tedious manual analysis of writing samples and careful consideration that the scores given by different evaluators had a high correlation, due to the subjective nature of the scoring.

These factors make in-clinic use prohibitive; as a result, these methods have received limited attention in follow-on work. 
The development of \emph{automated} and \emph{quantitative} metrics to analyze language complexity can potentially save several hours of research time to conduct similar linguistic studies to understand neurodegenerative disease and mental illness.
Several techniques devised in the NLP literature have been utilized to address the challenge of conducting quantitative analysis to replace traditionally subjective and task-dependent methods of measuring linguistic complexity.

\subsection{First Order Lexeme-Level Analysis}\label{subsec:lexical}
\subsubsection{Methods}
Automated first-order lexical analysis, \ie at the lexeme-level or word-level, can generate objective language metrics to provide valuable insight into cognitive function.
The most basic approaches treat a body of text as a \emph{bag of words}, meaning the ordering of words within the text is not considered.
This can be done by simply considering the frequency of usage of particular words and how they relate to a group of individuals.
Specialized tools, such as \emph{Linguistic Inquiry and Word Count} (LIWC)~\cite{tausczik2010psychological}, are often used to analyze the content and categorize the vocabulary within a text.
LIWC associates words in a text with categories associated with affective processes (\ie positive/negative emotions, anxiety, sadness, \etc), cognitive processes (\ie insight, certainty, causation, \etc), social processes (\ie friends, family, humans), the presence of dysfluencies (pauses, filler words, \etc), and many others.
The categorization of the lexicon allows for further tasks of interest, such as sentiment analysis based on the emotional categories.
The frequency of usage and other statistics of words from particular categories can lend insight to overall language production.

The concept of \emph{lexical diversity} refers to a measure of unique vocabulary usage.
The \emph{type-to-token ratio} ($\mathrm{TTR}$), given in Equation~\eqref{eq:ttr}, is a well-known measure of lexical diversity, in which the number of unique words (\emph{types}, $V$) are compared against the total number of words (\emph{tokens}, $N$). 
\begin{equation}\label{eq:ttr}
	\mathrm{TTR} = \frac{V}{N}
\end{equation}
However, $\mathrm{TTR}$ is negatively impacted for longer utterances, as the diversity of unique words typically plateaus as the number of total words increase.
The moving average type-to-token ratio ($\mathrm{MATTR}$) \cite{covingtonCuttingGordianKnot2010} is one method which aims to reduce the dependence on text length by considering $\mathrm{TTR}$ over a sliding window of the text.
This approach does not have a length-based bias, but is considerably more variable as the parameters are estimated on smaller speech samples. 
\emph{Brun\'et's Index} ($\mathrm{BI}$) \cite{brunet1978vocabulaire}, defined in Equation~\eqref{eq:brunet}, is another measure of lexical diversity that has a weaker dependence on text length, with a  smaller value indicating a greater degree of lexical diversity,
\begin{equation}\label{eq:brunet}
    \mathrm{BI} = N^{V^{-0.165}}.
\end{equation}
An alternative is also provided by \emph{Honor\'e's Statistic} ($\mathrm{HS}$) \cite{honore1979some}, defined in Equation~\eqref{eq:honore}, which emphasizes the use of words that are spoken only once (denoted by $V_1$),
\begin{equation}\label{eq:honore}
    \mathrm{HS} = 100 \log \frac{N}{1 - \nicefrac{V_1}{V}}.
\end{equation}
The exponential and logarithm in the $\mathrm{BI}$ and the $\mathrm{HS}$ reduce the dependence on the text length, while still using all samples to estimate the diversity measure, unlike the $\mathrm{MATTR}$.

Measures of \emph{lexical density}, which quantify the degree of information packaging within an utterance, may also be useful for cognitive assessment.
\emph{Content words}\footnote{Content words are also referred to as ``open-class'', meaning new words are often added and removed to this category of words as language changes over time.} (\ie nouns, verbs, adjectives, adverbs) tend to carry more information than \emph{function words}\footnote{Function words are also referred to as ``closed-class'' since words are rarely added to or removed from these categories.} (\eg prepositions, conjunctions, interjections, \etc).
These can be used to compute notions of \emph{content density} ($\mathrm{CD}$) in written or spoken language, given in Equation~\eqref{eq:content},
\begin{equation}\label{eq:content}
	\mathrm{CD} = \frac{\#~\text{of}~\mathrm{verbs}+\mathrm{nouns}+\mathrm{adjectives}+\mathrm{adverbs}}{N}.
\end{equation}
\emph{Part-of-speech (POS) tagging} of text samples is one way in which the word categories can be automatically determined; individual word tokens within a sentence are identified and labeled as the part-of-speech that they represent, typically from the Penn Treebank tagset \cite{marcusBuildingLargeAnnotated1993}.
Several automatic algorithms and available implementations exist for rule-based and statistical taggers, \ie using a \emph{hidden Markov model} (HMM) or \emph{maximum entropy Markov model} (MEMM) implementation to determine POS tags with a statistical sequence model \cite{jurafskySpeechLanguageProcessing2017}.
For example, the widely-used \emph{Stanford Tagger} \cite{toutanovaFeaturerichPartofspeechTagging2003} uses a bidirectional MEMM model to assign POS tags to samples of text.
Several notions of content density can be computed at the lexeme-level if POS tags can be automatically determined to reflect the role of each word in an utterance.
Examples of these include: the \emph{propositional density} ($P$-density), a measure of the number of expressed propositions (verbs, adjectives, adverbs, prepositions, and conjunctions) divided by the total number of words, and the \emph{content density}, which is a measure of the ratio of content words to function words~\cite{roarkSpokenLanguageDerived2011, fraserLinguisticFeaturesIdentify2015}.

\subsubsection{Clinical Applications}
Several studies have utilized first order lexical features to assess cognitive health by automated linguistic analysis.
The simplest bag-of-words analysis for vocabulary usage can often provide valuable insight in this regard.
For example, the work by Garrard \etal computed vocabulary statistics for participants with left- ($n=21$) and right-predominant ($n=11$) varieties of semantic dementia (SD) and, and compared them with language samples from healthy controls ($n=10$)~\cite{garrardMachineLearningApproaches2014}.
Classification accuracy of over $90\%$ was reached for categorizing the participants for two tasks: ($1$) participants with SD against the healthy control participants, and ($2$) classifying the left- and right-predominant variants of SD.
They used the concept of information gain to determine which word types were most useful in each classification problem.
Asgari \etal used the LIWC tool~\cite{tausczik2010psychological} to study the language of those with \emph{mild cognitive impairment} (MCI), often a precursor to Alzheimer's disease (AD)~\cite{asgariPredictingMildCognitive2017}.
The transcripts of unstructured conversation with the study's participants were analyzed with LIWC to generate a $68$-dimensional vector of word counts that fall within each of the $68$ subcategories in the LIWC lexicon.
They were able to achieve over $80\%$ classification accuracy by selecting LIWC categories that best represented the difference in the MCI and healthy control datasets.

Roark \etal considered a larger variety of speech and language features to detect MCI~\cite{roarkSpokenLanguageDerived2011}.
In this work, the authors compared the language of elderly healthy control participants and patients with MCI on the Wechsler Logical Memory I/II Test \cite{wechsler2008wechsler}, in which participants are tested on their ability to retell a short narrative that has been told them at different time points\footnote{Asked to retell the story immediately ($\mathrm{LM1}$) and after approximately $30$ minutes ($\mathrm{LM2}$)}.
Among the features considered included multiple measures of lexical density.
POS tagging was performed on the transcripts of clinical interviews of patients with MCI and healthy control participants.
Two measures of lexical density derived from the POS tags were the $P$-density and the content density.
In particular, the content density was a strong indicator of group differences between healthy controls and patients with MCI.
The automated language features were used in conjunction with speech features and clinical test scores to train a support vector machine (SVM) classifier that achieved good leave-pair-out cross validation results in classifying the two groups ($\mathrm{AUC} = 0.732,~0.703,~0.861$ when trained on language features, language features + speech features, and language~+ speech features + test scores, respectively)\footnote{Additional language and speech features will be discussed later}. 

Bucks \etal \cite{bucksAnalysisSpontaneousConversational2000} and Fraser \etal \cite{fraserLinguisticFeaturesIdentify2015} both used several first-order lexical features in their analysis of patients with AD.
In~\cite{bucksAnalysisSpontaneousConversational2000}, the authors successfully discriminated between a small sample of healthy older control participants ($n = 16$) and patients with AD ($n = 8$) using $\mathrm{TTR}$ (Equation~\eqref{eq:ttr}), $\mathrm{BI}$ (Equation~\eqref{eq:brunet}), and $\mathrm{HS}$ (Equation~\eqref{eq:honore}) as measures of lexical diversity or vocabulary richness.
They additionally considered the usage rates of other parts of speech (\ie nouns, pronouns, adjectives, verbs).
In particular, $\mathrm{TTR}$, $\mathrm{BI}$, verb-rate, and adjective-rate all indicated strong group differences between the participants with AD and healthy controls; the  groups could be classified with a cross-validation accuracy of $87.5\%$.
In~\cite{fraserLinguisticFeaturesIdentify2015}, Fraser \etal performed similar work using the \emph{DementiaBank}\footnote{\url{https://dementia.talkbank.org/access/}, Accessed August 20, 2019} database to obtain patient transcripts.
They additionally used other vocabulary-related features, such as frequency, familiarity, and imageability values for words in the transcripts.
This work was in turn based on a previous study~\cite{fraserAutomatedClassificationPrimary2012} in which similar features were extracted to study the language of participants with two different subtypes\footnote{Progressive nonfluent aphasia (PNFA) and semantic dementia (SD)} of primary progressive aphasia (PPA) and healthy control subjects.

Berisha \etal performed a longitudinal analysis of non-scripted press conference transcripts from U.S. Presidents Ronald Reagan (who was diagnosed with AD late in life) and George H.W. Bush (no such diagnosis) \cite{berishaTrackingDiscourseComplexity2015}.
Among the linguistic features that were tracked were the lexical diversity and lexical density for both presidents over several years worth of press conference transcripts. The study shows that the number of unique words used by Reagan over the period of his presidency steadily declined over time, while no such changes were seen for Bush. 
These declines predated his diagnosis of AD in 1994 by 6 years, suggesting that these computed lexical features may be useful in predicting the onset of AD pre-clinically. 
A related study examined the language in interview transcripts of professional American football players in the National Football League (NFL) \cite{berishaLongitudinalChangesLinguistic2017}, at high-risk for neurological damage in the form of chronic traumatic encephalopathy (CTE).
The study longitudinally measured $\mathrm{TTR}$~(Equation \ref{eq:ttr}) and $\mathrm{CD}$\footnote{The authors in \cite{berishaLongitudinalChangesLinguistic2017} refer to $\mathrm{CD}$ simply as ``lexical density'' ($\mathrm{LD}$)}~(Equation \ref{eq:content}) in interview transcripts of NFL players ($n = 10$) and NFL coaches/front office executives\footnote{Coaches and executives were limited to those who were not former players experiencing similar head trauma to serve as a control in the language study.} ($n = 18$).
Previous work has shown that $\mathrm{TTR}$ and $\mathrm{CD}$ are expected to increase or remain constant as healthy individuals age ~\cite{kemperAdultsDiariesChanges1990, kemperStructureVerbalAbilities2001, carlozziExaminationWechslerAdult2015}.
However, this study demonstrated clear longitudinal declines in both variables for the NFL players while showing the expected increase in both variables for coaches and executives in similar contexts.

\subsubsection{Advantages \& Disadvantages}
It is clear from the literature that first-order lexeme-level features, \ie those related to lexical diversity and density, are useful biomarkers for detecting the presence or predicting the onset of a variety of conditions, such as MCI, AD, CTE, and potentially several others.
POS tagging has several reliable and accurate implementations, and these features are simple and easy to compute.
Additionally, these linguistic measures are easily clinically interpretable for measuring cognitive-linguistic ability.

However, lexeme-level features are limited in what information they provide alone, and many of the previously discussed works used these features in conjunction with several other speech and language features to build their models for classification and prediction of disease onset.
Since these measures are based on counting particular word types and tokens, they tell us little about how individual lexical units interact with each other in a full sentence or phrase.
Additionally, measures of lexical diversity and lexical density provide little insight regarding semantic similarity between words within a sentence. 
For example, the words ``car'', ``vehicle'', and ``automobile'' are all counted as unique words, despite there being a clear semantic similarity between them\footnote{Note: lexical diversity is still a potentially useful measure in this case, as a diverse word choice may indicate higher cognitive function.}
In the following sections, we will discuss more complex language measures that aim to address these issues.

\subsection{Sentence-Level Syntactical Analysis} \label{subsec:syntax}
Generating free-flowing speech requires that we not only determine which words best convey an idea, but also to determine the order in which to sequence the words in forming sentences.
The complexity of the sentences we structure provides a great deal of insight into cognitive-linguistic health. 
In this section we provide an overview of various methods used to measure syntactic complexity as a proxy for cognitive health.

\subsubsection{Methods}
\begin{figure}
	\begin{subfigure}{\linewidth}
		\centering
		\includegraphics[width=0.85\linewidth]{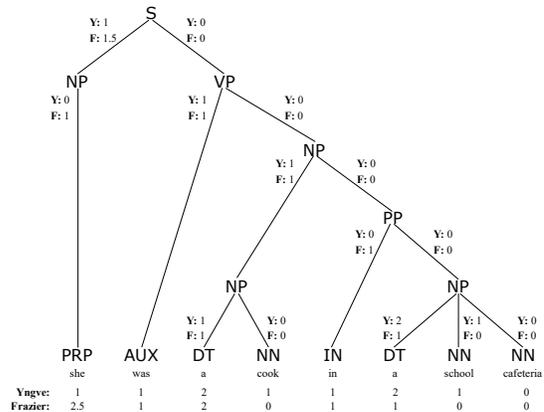}
		\caption{Constituency-based parsing of sample sentence (\ie top-down and left to right). In the diagram, \emph{S} = sentence, \emph{NP} = noun phrase, \emph{VP} = verb phrase, \emph{PP} = prepositional phrase, \emph{PRP} = personal pronoun, \emph{AUX} = auxiliary verb, \emph{DT} = determiner, \emph{NN} = noun, and \emph{IN} = preposition. The figure contains examples of both Yngve scoring ($\mathrm{Y}$) \cite{yngveModelHypothesisLanguage1960}, Frazier scoring ($\mathrm{F}$) \cite{frazierSyntacticComplexity1985} for each branch of the tree. At the bottom is the total score of each type for each word token in the sentence summed up to the root of the tree.}
		\label{fig:parsetree}
	\end{subfigure}

	\begin{subfigure}{\linewidth}
		\centering
		\includegraphics[width=0.85\linewidth]{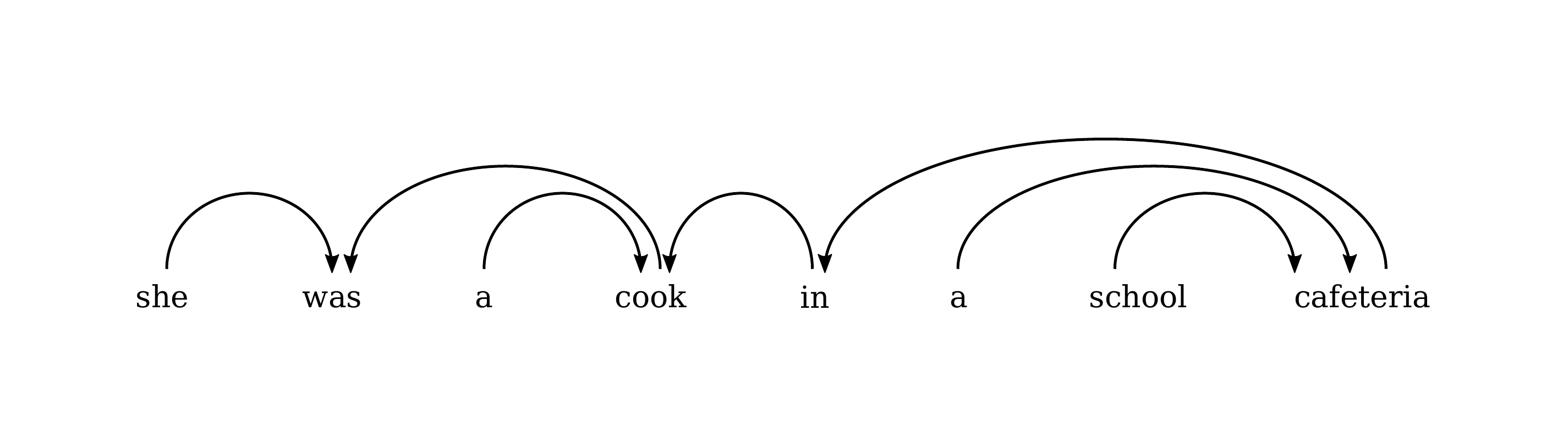}
		\caption{Dependency-based parsing of the same sample sentence. Lexical dependency distances can be computed.In this example, there are $7$ total links, a total lexical dependency distance of $11$, and an average distance of $\nicefrac{11}{7} = 1 \frac{4}{7}$.  Longer distances indicate greater linguistic complexity.}
		\label{fig:dependency}
	\end{subfigure}
	\centering
	\caption{(\subref{fig:parsetree}) A \emph{constituency-based} and (\subref{fig:dependency}) \emph{dependency-based} parsing of a simple sentence. Both adapted from~\protect\cite{roarkSpokenLanguageDerived2011}.}
\end{figure}

The ordering of words in sentences and sentences in paragraphs can also provide important insight into cognitive function. 
Many easy-to-compute and common structural metrics of language include the mean length of a clause, mean length of sentence, ratio of number of clauses to number of sentences, and other related statistics~\cite{fraserLinguisticFeaturesIdentify2015}.
Additionally, several more complicated methods for syntactical analysis of natural language can also be used to gain better insight for assessing linguistic complexity and cognitive health.

A commonly used technique involves the \emph{parsing} of naturally produced language based on language-dependent syntactical and grammatical rules.
A \emph{constituency-based parse tree} is generated to decompose a sentence or phrase into lexical units or tokens.
In English, for example, sentences are read left to right and are often parsed this way.
An example of a common constituency-based left to right parse tree can be seen in Figure~\ref{fig:parsetree} for the sentence ``She was a cook in a school cafeteria'', adapted from \cite{roarkSpokenLanguageDerived2011}.
At the root node, the sentence is split into a \emph{noun phrase} (``she'') and a \emph{verb phrase} (``was a cook in a school cafeteria'').
Then, the phrases are further parsed into individual tokens with a grammatical assignment (nouns, verbs, determiners, \etc).
Simple sentences in the English language are often \emph{right-branching} when using constituency-based parse trees.
This means that the subject typically appears first and is followed by the verb, object, and other modifiers.
This is primarily the case for the sentence in Figure~\ref{fig:parsetree}.
By contrast, \emph{left-branching} sentences place verbs, objects, or modifiers before the main subject of a sentence \cite{bergStructureLanguageDynamic2009}.
Left-branching sentences are often cognitively more taxing as they involve more complex constructions that require a speaker to remember more information about the subject before the subject is explicitly mentioned. 
As a result, in English, the degree of left-branching within a particular parsing of a sentence can be used as a proxy for syntactic complexity.

Once a parsing method has been implemented, various measures of lexical and syntactical complexity can be computed for each sentence or phrase.
Yngve proposes one such method in~\cite{yngveModelHypothesisLanguage1960}.
Given the right-branching nature of simple English sentences, he proposes a measure of complexity based on the amount of left-branching in a given sentence.
At each node in the parse tree, the rightmost branch is given a \emph{score} of $0$.
Then, each branch to the left of it is given a score that is incremented by $1$ when moving from right to left at a given node.
The score for each token is the sum of scores up all branches to the root of the tree.
An alternative scoring scheme for the same parse tree structure was proposed by Frazier \cite{frazierSyntacticComplexity1985}.
He notes that \emph{embedded clauses} within a sentence are an additional modifier that can increase the complexity of the syntactical construction of that sentence.
Therefore, just as with left-branching language, the speaker or listener would need to retain more information in order to properly convey or interpret the full sentence, respectively.
Frazier's scoring method emphasizes the use of embedded clauses when evaluating the syntactic complexity.
The scores are assigned to each lexeme as in Yngve's scoring, but they are summed up to the root of the tree or the lowest node that is not the leftmost child of its parent node.
Examples of both Yngve and Frazier scoring can be seen in Figure~\ref{fig:parsetree}.

Another type of syntactical parsing of a sentence is known as \emph{dependency parsing}, in which all nodes are treated as terminal nodes (no phrase categories such as \emph{verb phrase} or \emph{noun phrase}) \cite{magermanStatisticalDecisiontreeModels1995}.
A dependency-based parse tree aims to map the dependency of each word in a sentence or phrase to another within the same utterance.
Methods proposed by Lin~\cite{linStructuralComplexityNatural1996a} and Gibson~\cite{gibsonLinguisticComplexityLocality1998a} provide some ways by which the lexical dependency distances can be determined.
The general idea behind these methods is that longer lexical dependency distances within a sentence indicate a more complex linguistic structure, as the speaker and listener must remember more information about the dependencies between words in a sentence.
An example of the same sentence is shown with a dependency-based parse tree in Figure~\ref{fig:dependency}, also adapted from~\cite{roarkSpokenLanguageDerived2011}.

Mota \etal also propose a graph-theoretic approach for analyzing language structure as a marker of cognitive-linguistic ability with the construction of \emph{speech graphs} \cite{motaSpeechGraphsProvide2012, motaGraphAnalysisDream2014}.
\begin{figure}
	\begin{subfigure}{\linewidth}
		\centering
		\includegraphics[width=0.8\linewidth]{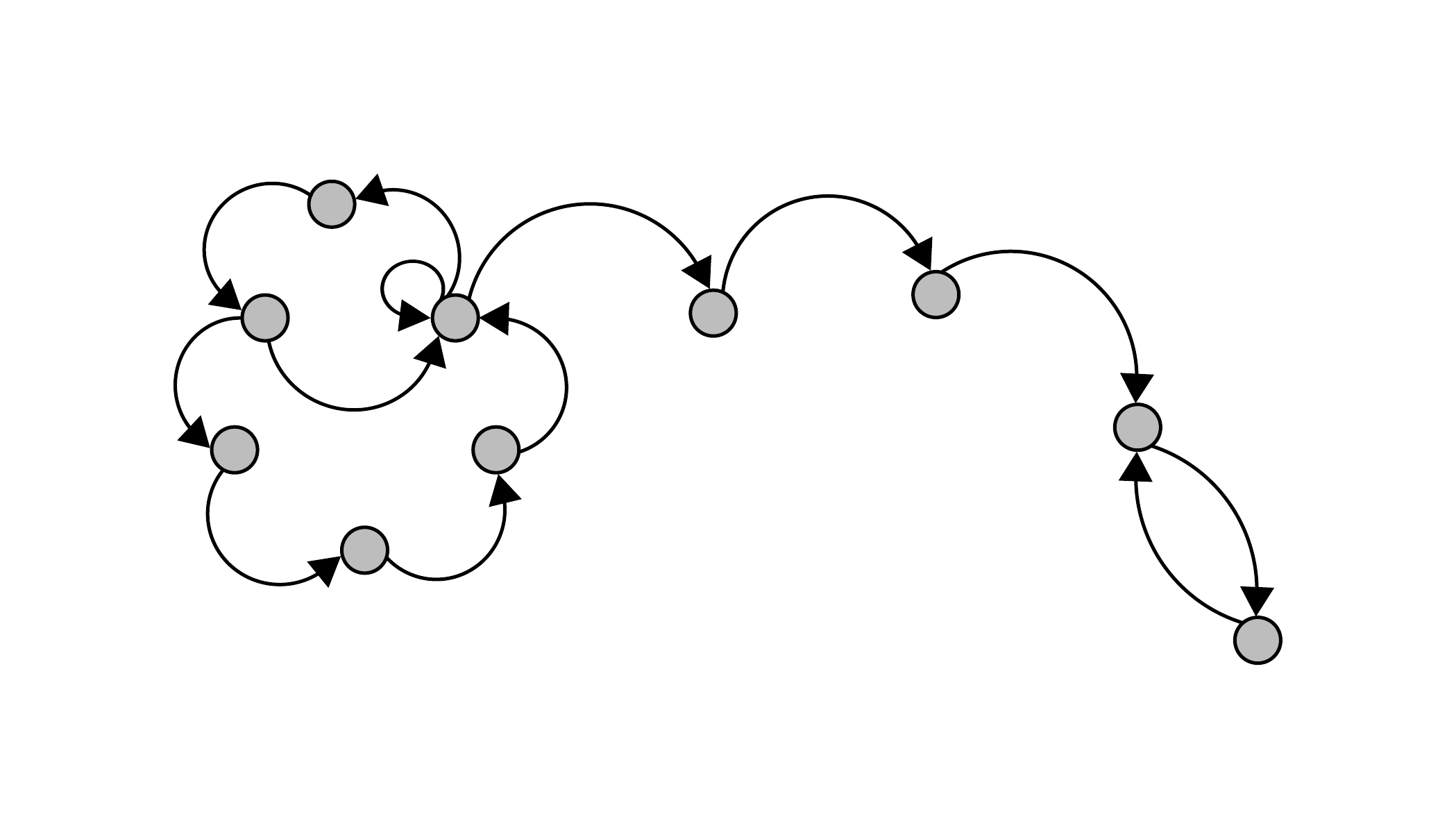}
		\caption{Sample speech-graph representation of a spoken utterance. Each of the circular nodes represents a lexical unit (\eg a single word) and the curved arrows represent edges which connect the relevant lexemes in the utterance. Attributes can be computed using the graph.}
		\label{fig:speechgraph}
	\end{subfigure}
	
	\begin{subfigure}{\linewidth}
		\centering
		\includegraphics[width=0.8\linewidth]{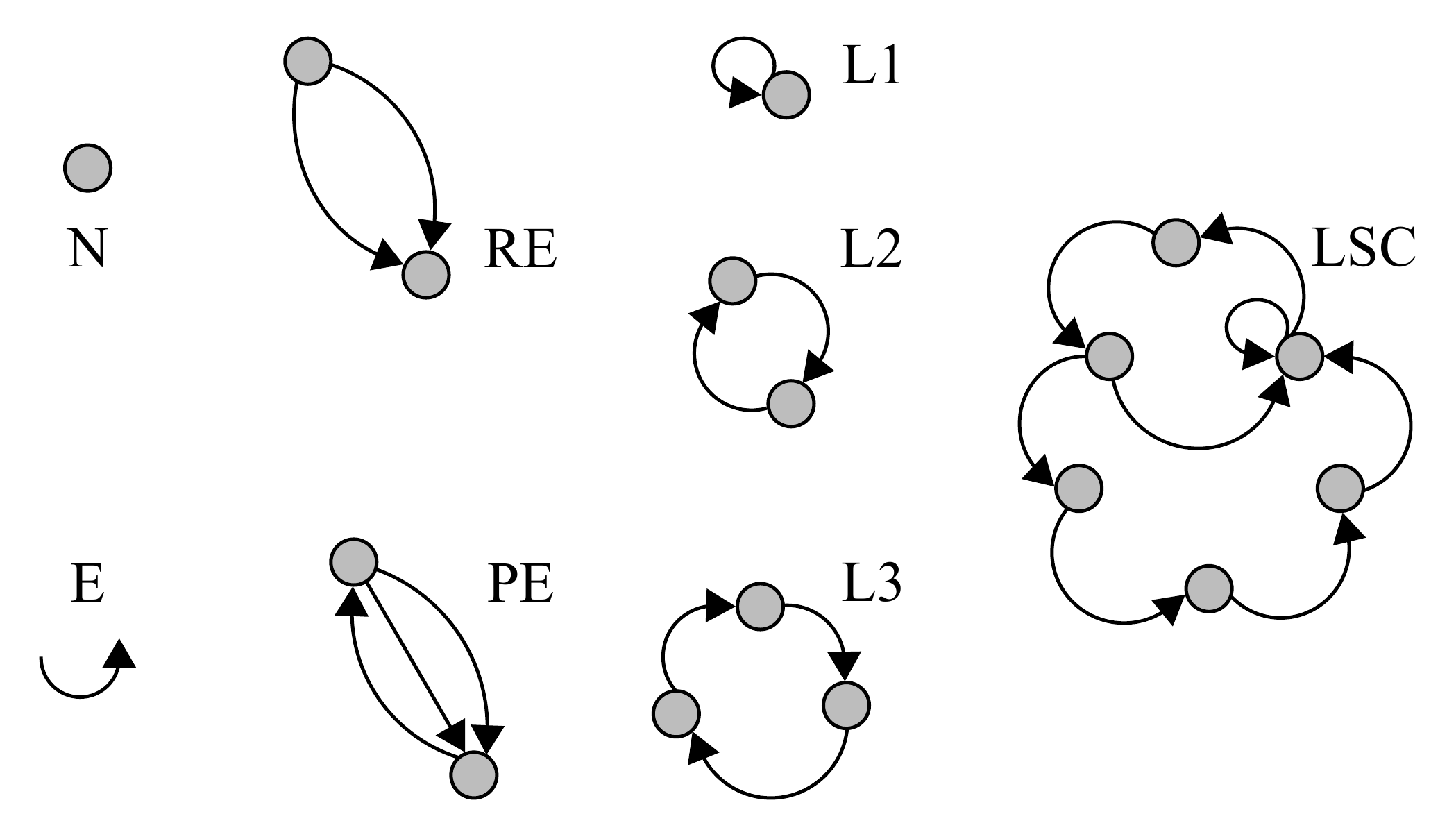}
		\caption{Examples of \emph{speech graph attributes} (SGAs). Examples include Nodes ($\mathrm{N}$), Edges ($\mathrm{E}$), Repeated Edges ($\mathrm{RE}$) in same direction, Parallel Edges ($\mathrm{PE}$), loops with $1$, $2$ and $3$ nodes ($\mathrm{L1}$, $\mathrm{L2}$, $\mathrm{L3}$), and the \emph{largest strongly connected component} ($\mathrm{LSC}$), \ie the portion of the total graph that can be reached from all others when considering the directionality of edges.}
		\label{fig:sga}
	\end{subfigure}
	\centering
	\caption{(\subref{fig:speechgraph}) A sample speech-graph for a complete spoken utterance. (\subref{fig:sga}) Example speech-graph attributes (SGAs). Both adapted from \protect\cite{motaGraphAnalysisDream2014}.}
\end{figure}
In this representation, the nodes are words that are connected to consecutive nodes in the sample text by edges representing lexical, grammatical, or semantic relationships between words in the text.
As an examples, for a speech graph based on words in an utterance, the spoken language is first transcribed and tokenized into individual lexemes, with each unique word by a graph node.
Directed edges then connect consecutive words\footnote{Speech graphs in some studies, \ie \cite{carrilloAutomatedSpeechAnalysis2016}, may use POS tags or other node structures}.
The researchers in this work suggest that structural graph features, \ie loop density, distance (number of nodes) between words of interest, \etc) serve as clinically relevant objective language measures that give insight into cognitive function.
An example speech-graph representation structure of an arbitrary utterance is seen in Figure~\ref{fig:speechgraph}.
The computed \emph{speech graph attributes} (SGAs) are the features which are extracted from these graphs, and some common ones visualized in Figure~\ref{fig:sga}.
The SGAs provide indirect measures of lexical diversity and syntactic complexity.
For example, $N$ is the number of unique words, $E$ is the total number of words, and repeated edges represent repeated words or phrases in text.

\subsubsection{Clinical Applications}
The structural aspects of spoken language have been shown to have clinical relevance for understanding medical conditions that affect cognitive-linguistic ability.
The previously mentioned work by Roark \etal also utilized several of the aforementioned methods to analyze the language of individuals with MCI and healthy control participants~\cite{roarkSpokenLanguageDerived2011}.
In addition to the lexeme-level features described in Section~\ref{subsec:lexical}, they also considered Yngve \cite{yngveModelHypothesisLanguage1960} and Frazier \cite{frazierSyntacticComplexity1985} scoring measures from constituency-based parsing of the transcripts of participant responses\footnote{Using the Charniak parser \cite{Charniak:2000:MP:974305.974323}}.
Mean, maximum, and average Yngve and Frazier scores were computed for each participant's language samples.
Roark \etal also used dependency parsing and computed lexical dependency distances, similar to the example in Figure~\ref{fig:dependency}.
Along with the lexical features and speech features, participants with MCI and healthy elderly control participants were classified successfully, as previously described in Section~\ref{subsec:lexical}.

The speech-graph approach is used by Mota \etal to study the language of patients with schizophrenia and bipolar disorder (mania)~\cite{motaSpeechGraphsProvide2012, motaGraphAnalysisDream2014}.
The researchers were able to identify structural features of the generated graphs (such as loop density, distance between words of interest, \etc) that serve as objective language measures containing clinically relevant information (\eg flight of thoughts, poverty of speech, \etc).
Using these features, the researchers were able to visualize and quantify concepts such as the \emph{logorrhea} (excessive wordiness and incoherence) associated with mania, evidenced by denser networks.
Similarly, the \emph{alogia} (poverty of speech) typical of schizophrenia was also visible in the generated speech-graph networks, as evidenced by a greater number of nodes per word and average total degree per node.
Control participants, participants with schizophrenia, and participants with mania were classified with over $90\%$ accuracy, significantly improving over traditional clinical measures, such as the Positive and Negative Syndrome Scale (PANSS) and Brief Psychiatric Rating Scale (BPRS)~\cite{motaSpeechGraphsProvide2012}.

\subsubsection{Advantages \& Disadvantages}
Consideration of sentence-level syntactical complexity offers several advantages that address some of the drawbacks of lexeme-level analysis.
As the work discussed here reveals, sentence structure metrics via syntactic parsing or speech-graph analysis offer powerful information in distinguishing healthy and clinical participants with schizophrenia, bipolar disorder/mania, mild cognitive impairment, and potentially several other conditions. 
Since sentence construction further taxes the cognitive-linguistic system beyond word finding, methods that capture sentence complexity provide more insight into the neurological health of the individual producing these utterances. 
This provides a multi-dimensional representation of cognitive-linguistics and allows for better characterization of different clinical conditions, as Mota \etal did with patients with schizophrenia and those with bipolar disorder/mania~\cite{motaSpeechGraphsProvide2012}.

However, while offering the ability to analyze more complex sentence structures, sentence-level syntactical analysis is also prone to increased complexity due to large range of implementation methodologies. 
For example, there are countless methods developed over the years for parsing language with different tools for measuring complexity relying on different algorithmic implementations of the language parsers, a widely studied topic in linguistic theory.
A thorough empirical evaluation of the various parsing methods is required to better characterize the performance of these methods in the context of clinical applications.

\subsection{Semantic Analysis}\label{subsec:semantics}
Cognitive function is also characterized by one's ability to convey organized and coherent thoughts through spoken or written language.
Here, we will cover some of the fundamental methods in NLP for measuring semantic coherence that have been used in clinical applications.

\subsubsection{Methods}
Semantic similarity in natural language is  typically measured computationally by \emph{embedding} text into a high-dimensional vector space that represents its semantic content.
Then, a notion of distance between vectors can be used to quantify semantic similarity or difference between the words or sentences represented by the vector embeddings.

Word embeddings are motivated by the \emph{distributional hypothesis} in linguistics, a concept proposed by English linguist John R. Firth who famously stated ``You shall know a word by the company it keeps''~\cite{haugenPapersLinguistics193419511958}, \ie that the inherent meaning of words is derived from their contextual usage in natural language.
One of the earliest developed word embedding methods is \emph{latent semantic analysis} (LSA)~\cite{landauerSolutionPlatoProblem1997}, in which word embeddings are determined by co-occurrence.
In LSA, each unit of text (such as a sentence, paragraph, document, \etc) within a corpus is modeled as a bag of words.

\begin{figure}
    \centering
    \includegraphics[width=0.9\linewidth]{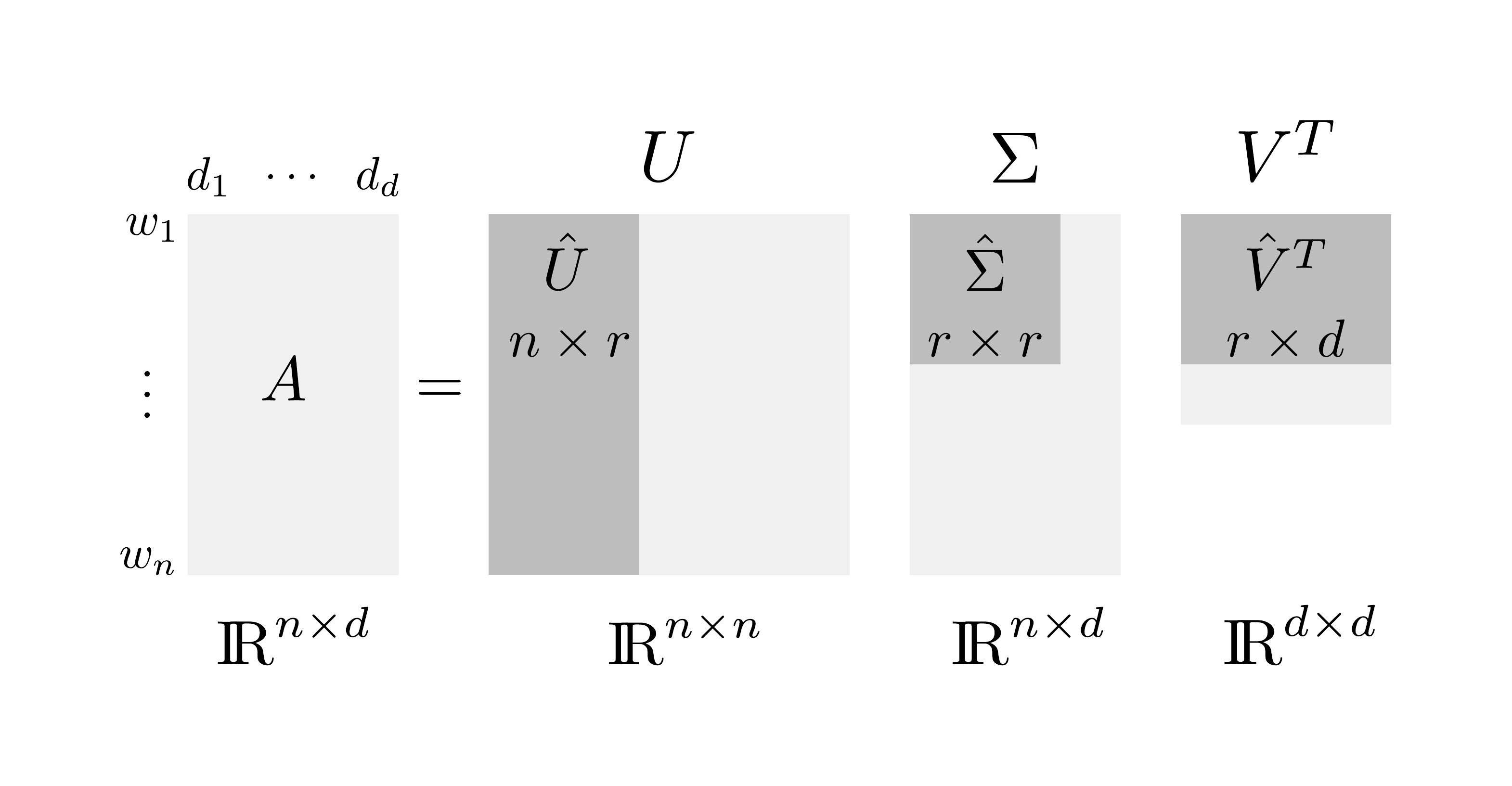}
    \caption{A visual representation of latent semantic analysis (LSA) by singular value decomposition (SVD).}
    \label{fig:lsa}
\end{figure}

As per Firth's hypothesis, the principal assumption of LSA is that words which occur together within a group of words will be semantically similar.
As seen in Figure~\ref{fig:lsa}, a matrix ($A$) is generated in which each row is a unique word in the text ($w_1, \ldots, w_n$) and each column represents a document or collection of text as described above ($d_1,\ldots d_d$).
The matrix entry values simply consist of the count of co-occurrence statistics, that is the number of times each word appears in each document.
Then a \emph{singular value decomposition} (SVD) is performed on $A$, such that $A = U \Sigma V^\mathrm{T}$. Here,  $U$ and $V$ are orthogonal matrices consisting of the left-singular and right-singular vectors (respectively) and $\Sigma$ is a rectangular diagonal matrix of singular values.
The diagonal elements of $\Sigma$ can be thought to represent semantic categories, the matrix $U$ represents a mapping from the words to the categories, and the matrix $V$ represents a mapping of documents to the same categories.
A subset of the $r$ most significant singular values is typically chosen, as shown by the matrix $\hat{\Sigma}$ in Figure~\ref{fig:lsa}.
This determines the dimension of the desired word embeddings (typically in the range of \textasciitilde$100$-$500$).
Similarly, the first $r$ columns of $U$ form the matrix $\hat{U}$ and the first $r$ rows of $V^\mathrm{T}$ form the matrix $\hat{V}^\mathrm{T}$.
The $r$-dimensional word embeddings for the $n$ unique words in the corpus are given by the resulting rows of the product $\hat{U} \hat{\Sigma}$.
Similarly, $r$-dimensional document embeddings can be generated by taking the $d$ columns of the product $\hat{\Sigma} \hat{V}^\mathrm{T}$.

\begin{figure}
    \centering
    \begin{subfigure}{0.75\linewidth}
        \caption{Continuous bag-of-words (CBOW)}
        \label{fig:w2v_cbow}
        \includegraphics[width=\linewidth]{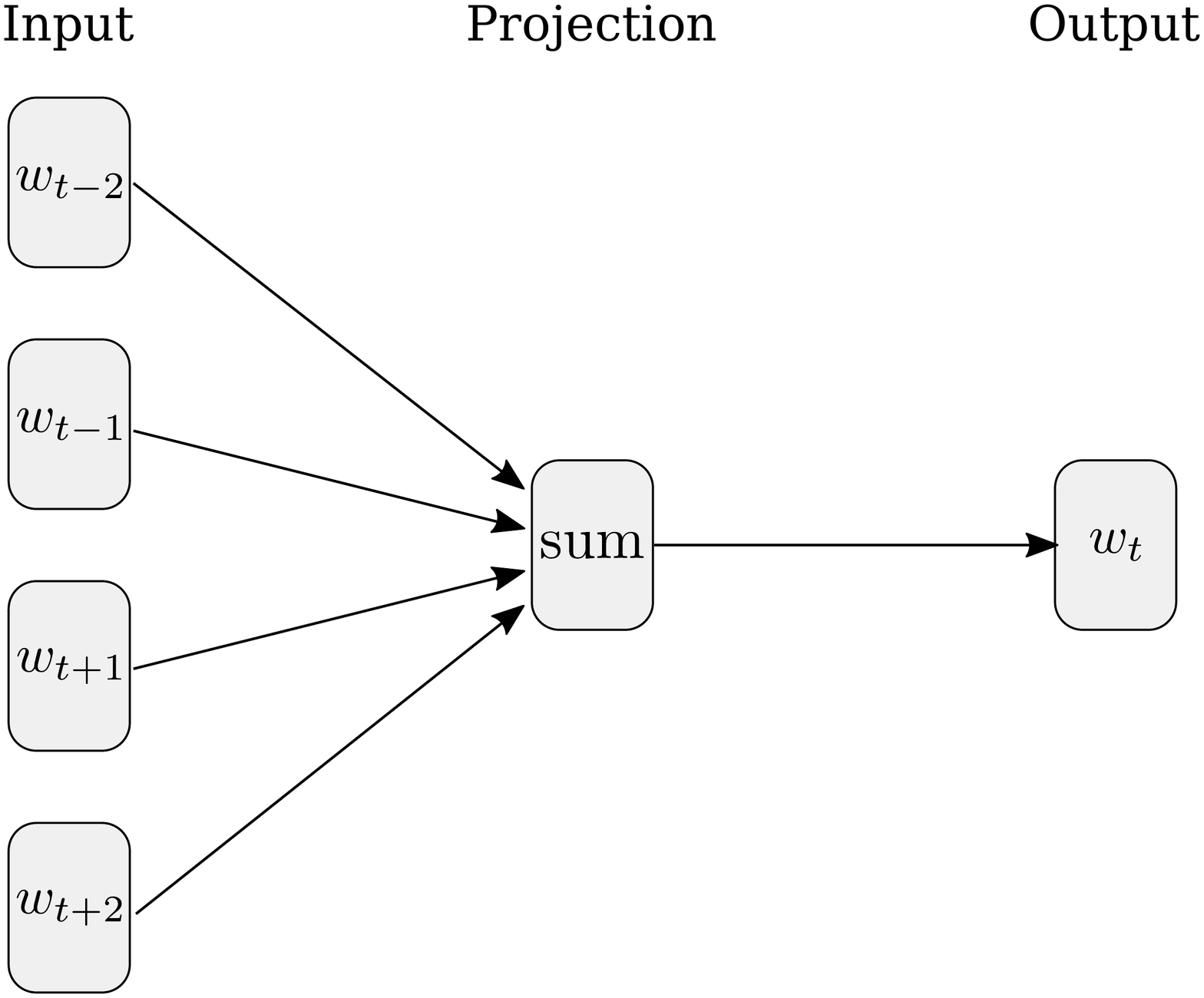}
    \end{subfigure}
    \begin{subfigure}{0.75\linewidth}
        \caption{Continuous skip-gram}
        \label{fig:w2v_sg}
        \includegraphics[width=\linewidth]{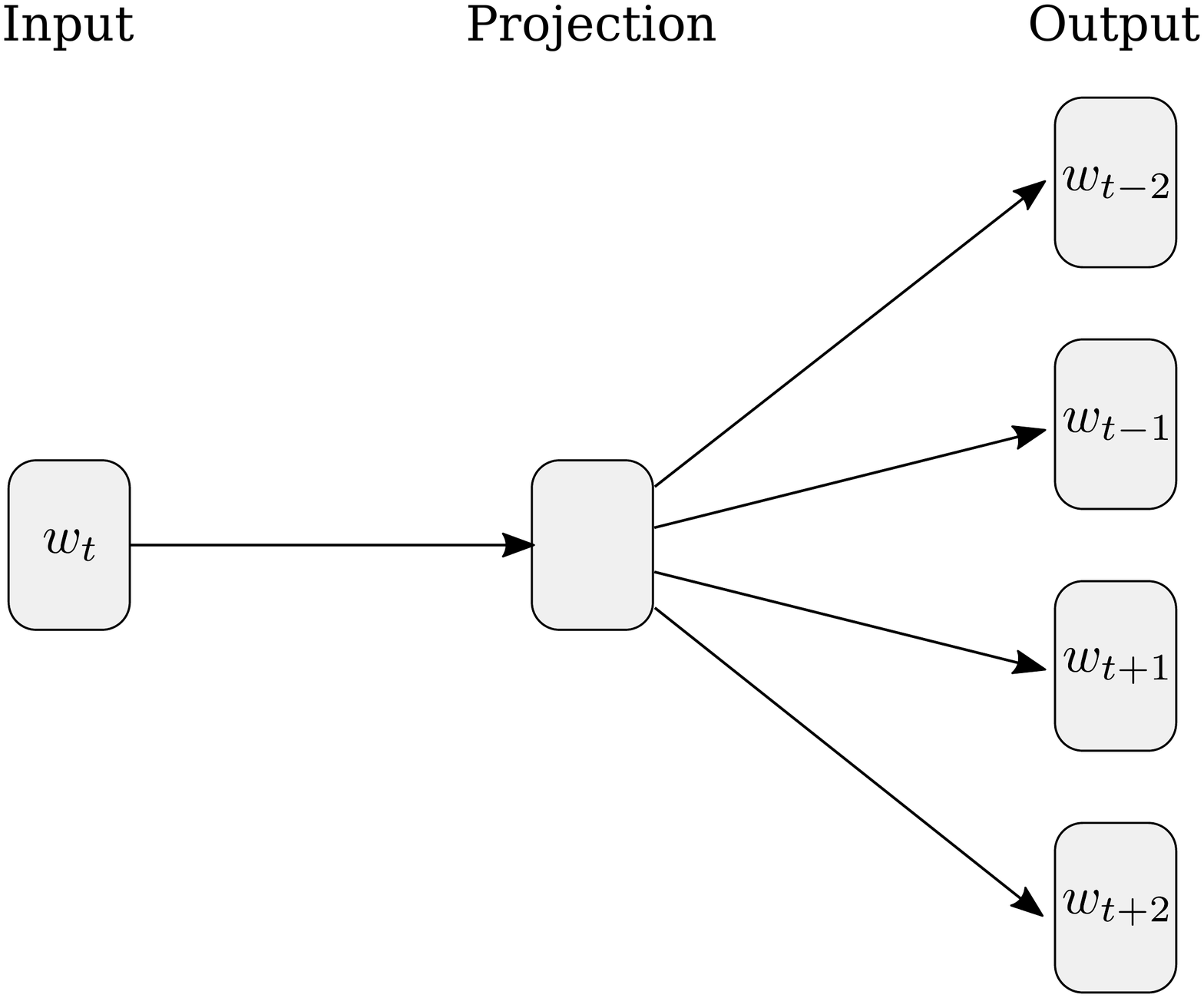}
    \end{subfigure}
    \caption{\emph{word2vec} model architectures proposed in \protect\cite{mikolovEfficientEstimationWord2013}. (\subref{fig:w2v_cbow}) In the CBOW model, the context words are inputs used to predict the center word. (\subref{fig:w2v_sg}) In the skip-gram model, the center word is used to predict the context words.}
    \label{fig:word2vec}
\end{figure}

In recent years, several new word embedding methods based on neural networks have gained popularity, such as \emph{word2vec}~\cite{mikolovEfficientEstimationWord2013} or \emph{GloVe}~\cite{penningtonGloveGlobalVectors2014}, which have shown improved performance over LSA for semantic modeling when sufficient training data is available~\cite{altszylerComparativeStudyLSA2016}.
As an example, we take a more detailed look at word2vec, proposed by Mikolov \etal, in which they present an efficient method for predicting word vectors based on very large corpora of text.
They present two versions of the word2vec algorithm, a \emph{continuous bag-of-words} (CBOW) model and \emph{continuous skip-gram} model, seen in Figure~\ref{fig:word2vec}.
At the input in both implementations, every word in a corpus of text is uniquely \emph{one-hot encoded}; \ie in a corpus of $V$ unique words, each word is uniquely encoded as a $V$-dimensional vector in which all elements are $0$ except for a single $1$.
In the CBOW implementation (Figure~\ref{fig:w2v_cbow}), the inputs are the context words in the particular neighborhood of a target center word, $w_t$. In the skip-gram implementation (Figure~\ref{fig:w2v_sg}), the input is the center word and the objective is to predict the context words at the output.
In both models, the latent hidden representation of dimension $N$ gives an embedding for the word represented by the one-hot encoded input word.
The training objective is to minimize the cross-entropy loss for the prediction outcomes.

There are several other methods for word embeddings, each relying on the distributional hypothesis and each with various advantages and disadvantages.
For example, LSA, \emph{word2vec} and \emph{GloVe} are simple to train and effective, but a major disadvantage is that they do not handle out-of-vocabulary (OOV) words or consider words with multiple unrelated meanings.
For example, the English word ``bark'' can refer to the bark of a dog or to the bark of a tree, but its vector representation would be an average representation, despite the drastically different usage in each context.
Some methods based on deep neural networks (DNNs), such as recurrent neural network (RNN) / long-short term memory (LSTM) networks (\eg ELMo~\cite{petersDeepContextualizedWord2018}) or transformer architectures (\eg BERT~\cite{devlinBERTPretrainingDeep2018}) utilize contextual information to generate embeddings for OOV words.

In addition to individual words, embeddings can also be learned at the sentence level.
The simplest forms of sentence embeddings involve unweighted averaging of LSA, \emph{word2vec}, \emph{GloVe}, or other embeddings.
Weighted averages can also be computed, such as by using \emph{term frequency-inverse document frequency} (tf-idf) generated weights or \emph{Smooth Inverse Frequency} (SIF)~\cite{aroraSimpleToughtoBeatBaseline2017}.
Others have found success learning sentence representations directly, such as in \emph{sent2vec}~\cite{pagliardiniUnsupervisedLearningSentence2017}.
Whole sentence encoders, such as \emph{InferSent}~\cite{conneauSupervisedLearningUniversal2017} and the \emph{Universal Sentence Encoder} (USE)~\cite{cer2018universal} offer the advantage of learning a full sentence encoding that considers word order within a sentence; \eg the sentences ``The man bites the dog'' and the ``The dog bites the man'' will each have different encodings though they contain the same words.

Once an embedding has been defined, a notion of semantic similarity or difference must also be defined.
Several notions of distance can be computed for vectors in high-dimensional space, such as Manhattan distance ($\ell_1$ norm), Euclidean distance ($\ell_2$ norm), or many others.
Empirically, the \emph{cosine similarity}  (cosine of the angle, $\theta$, between vectors) has been found to work well in defining semantic similarity between word and sentence vectors of many types.
Cosine similarity can be computed using Equation~\eqref{eq:cosine} for vectors $\mathbf{w}_1$ and $\mathbf{w}_2$.
\begin{equation}\label{eq:cosine}
    \mathrm{CosSim}(\mathbf{w}_1, \mathbf{w}_2) = \cos \theta = \frac{\mathbf{w}_1^\mathrm{T} \mathbf{w}_2}{\norm{\mathbf{w}_1}_2 \norm{\mathbf{w}_2}_2}
\end{equation}

In addition to word and sentence embedding semantic similarity measures, techniques such as \emph{topic modeling} and \emph{semantic role labeling} have also gained recently popularity in NLP and its applications to clinical language samples.
\emph{Latent dirichlet analysis} (LDA) is one such statistical topic modeling method which can be used to identify overarching themes in samples of text~\cite{bleiLatentDirichletAllocation2003}.
Other studies have utilized \emph{semantic role labeling}, a probabilistic technique which automatically attempts to identify the semantic role a particularly entity plays in a sentence~\cite{dasProbabilisticFramesemanticParsing2010}.

\subsubsection{Clinical Application}
Many forms of mental illness can result in a condition known as \emph{formal thought disorder} (FTD), which impairs an individual's ability to produce semantically coherent language.
FTD is most commonly associated with schizophrenia but is often present in other forms of mental illness such as mania, depression, and several others~\cite{colmanDictionaryPsychology2015, yudofskyAmericanPsychiatricPublishing2002}.
Some common symptoms include poverty of speech (\emph{alogia}), derailment of speech, and semantically incoherent speech (\emph{word salad})~\cite{yudofskyAmericanPsychiatricPublishing2002, videbeck2010psychiatric}.
Language metrics that track semantic coherence are potentially useful in clinical applications, such as measuring the coherence of language as it relates to FTD in schizophrenia.
One of the first studies to demonstrate this was conducted by Elvev\r{a}g \etal~\cite{elvevagQuantifyingIncoherenceSpeech2007}.
The language of patients with varying degrees of FTD (rated by standard clinical scales) was compared with a group of healthy control participants.
The experimental tasks consisted of single word associations, verbal fluency (naming as many words as possible within a specific category), long interview responses (\textasciitilde1-2 minutes per response), and storytelling.
LSA was utilized to embed the word tokens in the transcripts.
The semantic coherence in each task was computed as follows:
\begin{itemize}
    \item \emph{Word Associations}: Cosine similarity between cue word and response word, with an average coherence score for each participant
    \item \emph{Verbal Fluency}: Cosine similarity between first and second word, second and third word, etc. were computed, with an average coherence score computed per participant
    \item \emph{Interviews}: Cosine similarity was computed between the question and participant responses. An average word vector was computed for the prompt question from the interviewer. Then a \emph{moving window} (of size 2-6 words) for the participant response was used to average all the word vectors within the window and compute a cosine similarity between the question and response. The window was moved over the entire participant response and a new cosine similarity was computed between the question and response window until reaching the end of the response. This method tracks how the cosine similarity behaves as the participant response goes farther from the question, with the expectation that the response would be more tangential over time with decreased coherence as the participant moves farther from the question. A regression line was fit for each participant to measure the change in cosine similarity coherence over time, and the slope of the line was computed to measure the tangentiality of the response per participant.
    \item \emph{Storytelling}: Cosine similarity of the participant's response was compared to the centroid participant response for all narrative utterances of the same story. This was used to predict the clinical rating for thought disordered language samples when asked to tell the same story.
\end{itemize}
They demonstrated that the control participants had higher coherence scores compared to the FTD groups across all tasks.

In a more recent study, predictive features of language for the onset of psychosis were studied by Bedi \etal~\cite{bediAutomatedAnalysisFree2015}.
Open-ended narrative-like interview transcripts of young individuals who were determined to be at \emph{clinical high-risk} (CHR) for psychosis were collected and analyzed to predict which individuals would eventually develop psychosis.
Participants were tracked and interviewed over a period of two and a half years.
In this study, LSA was again used to generate word embeddings.
An average vector for each phrase was computed, and a cosine-similarity measure was computed to measure the semantic coherence between consecutive phrases (\emph{first-order coherence}) and every other phrase (\emph{second-order coherence}).

A distribution of the first and second-order coherence scores (cosine similarities) was compiled for each participant, and several statistics were computed based on the distribution of coherence scores, \eg maximum, minimum, standard deviation, mean, median, $10^{\text{th}}$ percentile, and  $90^{\text{th}}$ percentile.
Each of these statistics was considered as a separating feature between the clinical and control samples.
In addition to the semantic analysis, POS-tagging was performed to compute the frequency of use of each part-of-speech to obtain information about the structure of each participant's naturally-produced language.
The language features with the best predictive power in the classifier were the minimum coherence between consecutive phrases for each participant (maximum discontinuity) and the frequency of use of determiners (normalized by sentence length).
This initial study only had 34 participants total (only 5 CHR+ participants) and was intended as a  proof-of-principle exploration. 
In an expansion of this work, Corcoran \etal trained their classifier using two larger datasets, in which one group of participants was questioned with a prompt-based protocol and another group of participants was given a narrative protocol in which they were required to provide longer answers (similar to the previous work)~\cite{corcoranPredictionPsychosisProtocols2018}.
They note that the first and second-order coherence metrics collected in the previous study were useful for determining semantic coherence with the narrative-style interview transcripts with longer responses.
However, for the shorter prompt-based responses (often under 20 words), it is often difficult to obtain these metrics.
Therefore, coherence was-computed on the \emph{word-level} rather than phrase-level by computing the cosine similarity between word embeddings within a response with an inter-word distance of $k$, with $k$ ranging from 5 to 8.
As before, typical statistics were computed on the coherence values obtained for each participant response (maximum, minimum, mean, median, 90\textsuperscript{th} percentile, 10\textsuperscript{th} percentile, \etc).
They were able to successfully predict the onset of schizophrenia by discriminating the speech of healthy controls and those with early onset schizophrenia with $ \sim$ 80\% accuracy.

Other studies make use of a variety of linguistic features to predict the presence of clinical conditions.
For example, Kayi \etal identified predictive linguistic features of schizophrenia by analyzing laboratory writing samples of patients and controls for their semantic, syntactic, and pragmatic (sentimental) content~\cite{kayiPredictiveLinguisticFeatures2017}.
A second dataset of social media messages from self-reporting individuals with schizophrenia over the Twitter API was also evaluated for the same types of content.
The semantic content of the language was quantified by three methods:
First, semantic role labeling was performed using the Semafor tool~\cite{dasProbabilisticFramesemanticParsing2010} to identify the role of individual words within a sentence or phrase
Then, LDA was used to identify overarching themes that separated the clinical and control writing samples~\cite{bleiLatentDirichletAllocation2003}.
LDA identifies topics in the text and also identifies the top vocabulary used in each topic.
Finally, clusters of word embeddings within the writing were generated using the $k$-means algorithm and \emph{GloVe} word vector embeddings~\cite{penningtonGloveGlobalVectors2014}.
The frequency of each cluster was computed per document by checking the use of each word of the document in each cluster.
The syntactic features used in this study again were obtained by computing the frequency of use of parts of speech (found by POS tagging) and by generating parse trees, using tools optimized for the corpus.
Lastly, pragmatic features were found by performing \emph{sentiment analysis} to classify the sentiment of the writing samples into distinct groups (\emph{very negative}, \emph{negative}, \emph{neutral}, \emph{positive}, \emph{very positive}).
They successfully showed a distinct set of predictive features that could accurately separate participants with schizophrenia from healthy controls in all of the language analysis categories.
However, when using a combination of features and various machine learning classifiers (random forest and support vector machine), they found that utilizing a combination of the semantic and pragmatic features led to the most promising accuracy (81.7\%) in classification of control participants and those with schizophrenia.
The limited availability of language data in schizophrenia is always a difficult challenge, so another study by Mitchell \etal analyzed publicly available social media (Twitter) posts by self-identifying individuals with schizophrenia using LDA, LIWC generated-vectors, and various clustering techniques to show statistically significant differences in their language patterns when compared to general users~\cite{mitchell2015quantifying}.

Another vector-space topic modeling approach was developed by Yancheva and Rudzicz for analyzing transcripts of picture description tasks for participants with AD and healthy controls~\cite{yanchevaVectorspaceTopicModels2016}.
They propose a general method for generating \emph{information content units} (ICUs), or topics of interest, from common images used in clinical description task evaluations, \ie the famous \emph{Cookie Theft} picture with reference speech samples~\cite{goodglassAssessmentAphasiaRelated1983}.
The generated ICUs were compared with human-supplied ICUs from common usage in clinical practice, and most of the categories exhibited a close match.
The study found that participants with AD and healthy controls were likely to discuss the same topics, but those with AD had wider topic clusters with more irrelevant directions.
Additionally, they were able to find a small set of generated ICUs that had slightly better classification performance than a much larger set of human selected ICUs for the same task, with  $ \sim 80\%$ accuracy. 
Related work by Hern\'andez-Dom\'inguez \etal took a similar approach to generate a population-specific set of categories for participants with AD ($n=257$), MCI ($n=43$), and healthy controls ($n=217$)~\cite{hernandez-dominguezComputerbasedEvaluationAlzheimer2018}.
The resulting features were significantly correlated with severity as assessed by the MMSE, and classification performance was characterized by the receiver operating characteristic (ROC) area under curve (AUC) performance of $\mathrm{AUC} \approx 0.76$ for all three groups.

\subsubsection{Advantages and Disadvantages}
While these studies have been successful in measuring the semantic coherence of language as it relates to thought disorders, there are several limitations.
Recent work by Iter \etal identifies and attempts to address some of these shortcomings when measuring semantic coherence for FTD in schizophrenia~\cite{iter2018automatic}.
Interviews with a small sample of patients were collected and just the participant responses (of \textasciitilde300 words each) were analyzed for their semantic content.
They noted that when using the \emph{tangentiality model} of semantic coherence (i.e. regression of the coherence over time with the sliding window) of Elvev\r{a}g \etal~\cite{elvevagQuantifyingIncoherenceSpeech2007} and the \emph{incoherence model} of semantic coherence of Bedi \etal~\cite{bediAutomatedAnalysisFree2015}, they were unable to convincingly separate their clinical and control participants based on language analysis.
One reason for this was due to the presence of verbal fillers, such as "um" or "uh" and many \emph{stop words} without meaningful semantic content.
Another reason is that longer sentences (or long moving windows) tend to be scored as more coherent due to a larger overlap of words.
The third reason they identified (but did not address) is that repetitive sentences and phrases would be scored as highly coherent, even though repetition of ideas is common in FTD and should be scored negatively.
The authors proposed a series of improvements to address some of these limitations. 
However, the sample sizes in this study were small (9 clinical participants and 5 control participants).

Another issue with semantic coherence computation in clinical practice is difficulty with interpretability of computed metrics; for example, the cosine similarity between high dimensional word vectors is a somewhat abstract concept which is difficult for most to visualize.
Recent work~\cite{Voleti2019} attempted to address this issue by computing semantic coherence measures (using \emph{word2vec}, \emph{InferSent}, and SIF embeddings), lexical density and diversity measures, and syntactic complexity measures as they relate to the language of patients with schizophrenia, patients with bipolar disorder, and healthy controls undergoing a validated clinical social skills assessment~\cite{pattersonSocialSkillsPerformance2001}.
Linear regression was used to determine a subset of language features across all categories that could effectively model the scores assigned by clinicians during the social skills performance assessment, in which participants were required to act out various role-playing conversational scenes with clinical assessors scored for cognitive performance.
Then, these features were used to train simple binary classifiers (both na\"ive Bayes and logistic regression), for which leave-one-out cross-validation was used to determine their effectiveness at classifying groups of interest.
For classifying clinical (patients with schizophrenia and bipolar I disorder) participants and healthy control participants, the selected feature subset achieved ROC curve AUC performance of $\mathrm{AUC} \approx 0.90$; for classifying within the clinical group (to separate participants with schizophrenia and bipolar disorder), the classifier performance achieved $\mathrm{AUC} \approx 0.80$.
\section{Measuring Cognitive and Thought Disorders with Speech Signal Processing}\label{sec:speech}
While cognitive-linguistic health is more directly observed through analysis of complex language production, additional information can be derived by speech signal analysis of individuals with cognitive impairments or thought disorders.
This is because the acoustic stream is the physical manifestation of the cognitive-linguistic processing that has gone into creating the message being conveyed, in near real-time.
In this way, pauses during speech can be associated with difficulty in lexical retrieval (word-finding difficulties) or with extra processing time needed for message formulation.
Pressed speech, that which is rapidly produced without insertions of natural pauses, can be associated with mania and ``flight of thoughts''. 
Conversely, reductions in the rhythmic and melodic variations in speech may be indicative of changes in mood.

The information derived from the speech signal is used alone or in conjunction with many of the previously described methods to assess cognitive-linguistic health. This is either done directly by measuring different aspects of speech production including prosody, articulation, or vocal quality; or is done as a pre-processing step by using 
automatic speech recognition (ASR) for transcription of speech samples for follow-on linguistic analysis. 

In this section, we will review how various signal processing methods are used to extract clinically-relevant insight from an individual's speech samples for additional insight into detection of disorders that affect cognition and thought. Referring back to Fig. 2, these include features extracted from vocal fold vibration (source), movement of the articulators (filter), and the overall rhythm of the speech signal (prosody). 

\subsection{Methods}\label{subsec:speech_methods}

\subsubsection{Prosodic features}
Prosody refers to the rhythm and melody of speech. Examples of computable temporal prosodic features from recorded speech signals include the duration of voiced segments, duration of silent segments, loudness, measures of periodicity, fundamental frequency ($F_0$), and many other similar features \cite{roarkSpokenLanguageDerived2011, konigAutomaticSpeechAnalysis2015}.
These measures can indicate irregularities in the rhythm and timing of speech.
Additionally, nonverbal speech cues, \eg counting the number of interruptions, interjections, natural turns, and response times can also indicate identifying features of irregular speech patterns~\cite{tahirNonverbalSpeechAnalysis2016}.

\subsubsection{Articulation features}
Several spectral features that capture movement of the articulators have been used in the clinical speech literature to measure the acoustic manifestation of the cognitive-linguistic deficits discussed in Section 2. These include computing statistics related to the presence of additional formant harmonic frequencies, \ie $F_1$, $F_2$, and $F_3$, computing formant trajectories over time~\cite{horwitz-martinRelationAutomaticallyExtracted2016}, or computing the vowel space area~\cite{sandovalAutomaticAssessmentVowel2013}.  
The \emph{spectral centroid} can also be computed for each frame of speech signal that is analyzed~\cite{peeters2004large}.
The spectral centroid is essentially the center of mass for the frequency spectrum of a signal, and relates to the ``brightness'' or timbre of the perceived sound for audio.

Time-frequency signal processing techniques are also commonly used since acoustic speech signals are highly non-stationary.
For example, computation of the mel-frequency cepstral coefficients (MFCC) with the mel scale filterbank provides a compressed and whitened spectral representation of the speech ~\cite{davisComparisonParametricRepresentations1980}.
These features are often used as inputs into an automatic speech recognition (ASR) system, but can also be monitored over time to identify irregularities in speech due to cognitive or thought disorders.
As an example, common statistical features such as the mean, variance, skewness, and kurtosis of the MFCCs over time can be tracked for identification of irregularities between healthy individuals and those with some cognitive or thought disorders \cite{fraserLinguisticFeaturesIdentify2015}.

\subsubsection{Vocal quality features}
There is evidence that there are vocal quality changes associated with cognitive disorders~\cite{hailstoneVoiceProcessingDementia2011}. 
These can be measured from the speech signal by isolating the \emph{source} of speech production, involving the flow of air through the lungs and glottis and affecting perceptible voice quality.
Voice quality measures that have previously been used in the context of cognitive and thought disorders include:
\begin{itemize}
	\item \emph{jitter}: small variations in glottal pulse \emph{timing} during voiced speech
	\item \emph{shimmer}: small variations in glottal pulse \emph{amplitude} during voiced speech
	\item \emph{harmonic-to-noise ratio} (HNR): the ratio of formant harmonics to inharmonic spectral peaks, \ie those that are not whole number multiples of $F_0$
\end{itemize}
These features alone are often difficult to consistently compute and interpret, but can provide insight for the diagnosis and characterization of certain clinical conditions.

\subsubsection{Automatic Speech Recognition}
Recent improvements in ASR and in tools for easily implementing ASR systems have made possible the use of these systems in clinical speech analysis.
This is most commonly done by using ASR in place of manual transcription for the extraction of linguistic features (\ie features covered in Section~\ref{sec:nlp}); however, this is often more error prone with regard to incorrect word substitutions, unintended insertions, or unintended deletions in the automatically generated transcript.
The \emph{word error rate} ($\mathrm{WER}$) for an utterance of $N$ words is given in Equation~\eqref{eq:wer}, 
\begin{equation}\label{eq:wer}
	\mathrm{WER} = \frac{\mathrm{\#~of~insertions} + \mathrm{deletions} + \mathrm{substitutions}}{N},
\end{equation}
and is a typical statistic used to evaluate the performance of an ASR system.
It is often more difficult to maintain high accuracy (low $\mathrm{WER}$) for ASR with pathological speech samples, as the relative dearth of this data makes it difficult to train reliable ASR models optimized for this task.
Other studies have also made use of ASR for paralinguistic feature extraction, such as the automated detection of filled pauses, natural turns, interjections, \etc~
Understanding the effects of ASR errors on downstream NLP tasks is an important area to address in which the current work is limited.
Some recent attempts have been made to simulate ASR errors on text datasets and evaluate their effects on downstream tasks~\cite{stuttleFrameworkDialogueData2004,simonnetSimulatingASRErrors2018, voletiInvestigatingEffectsWord2019}.
These potentially have future applications in language models that can analyze noisy datasets with ASR errors in clinical practice.

\subsection{Clinical Applications}~\label{subsec:speech_applications}
\subsubsection{Acoustic analysis}
Disorders such as PPA, MCI, AD, and other forms of dementia are associated with a general slowing of thoughts in affected individuals.
This has been shown to have detectable effects on speech production through acoustic analysis.
In a study by K\"onig \etal, healthy controls and participants with MCI and AD were recorded as they were asked to perform various tasks, such as counting backwards, image description, sentence repeating, and verbal fluency testing~\cite{konigAutomaticSpeechAnalysis2015}.
Temporal prosodic features such as the duration of voiced segments, silent segments, periodic segments, and aperiodic segments were all computed.
Then, the ratio of the mean durations of voiced segments to silent segments were also computed as features to express the continuity of speech in the study's participants.
As expected, it was shown that healthy control participants showed greater continuity in these metrics when compared to those with MCI or AD. 
These quantifiable alterations of speech in individuals with MCI and AD allowed the researchers to successfully separate patients with AD from healthy controls (approx. $87\%$ accuracy), patients with MCI from controls (approx. $80\%$ accuracy), and patients with MCI from patients with AD (approx. $80\%$ accuracy).
L\'opez-de-Ipi\~na \etal conducted another study in which acoustic features (related to prosody, spectral analysis, and features with emotional content) were extracted from spontaneous speech samples to classify participants with AD at different stages (early, intermediate, and advanced)~\cite{lopez-de-ipinaAutomaticDiagnosisAlzheimer2015}.
Among the computed prosodic features were the mean, median, and variance for durations of voiced and voiceless segments. 
Short-time energy computations were also computed for the collected samples in the time-domain.
In the frequency-domain, the spectral centroid was determined for each speech sample.
The authors also claim that features such as the contour of $F_0$ and source features like shimmer, jitter, and noise-harmonics ratio contain emotional content that can be useful in the automatic AD diagnosis.
Lastly, they propose a new feature, which they term \emph{emotional temperature} ($\mathrm{ET}$), which is a normalized (independent of participant) measure ranging from $0$-$100$ based on several of prosodic and paralanguistic features that were previously mentioned\footnote{The example in \cite{lopez-de-ipinaAutomaticDiagnosisAlzheimer2015} shows that a typical $\mathrm{ET}$ value is approx. $95$ for healthy control participants and approx. $50$ for those with AD}.
The study revealed several interesting findings.
First, the spontaneous speech analysis indicated that participants with AD exhibited higher proportions of voiceless speech and lower proportions of voiced speech, indicating a loss of fluency and shortening of fluent speech segments for those with AD.
While classification accuracy was good when using a set of prosodic speech features, they noted that accuracy improved when the emotional features (\ie the proposed $\mathrm{ET}$ metric) were used\footnote{see Figure~$9$ in~\cite{lopez-de-ipinaAutomaticDiagnosisAlzheimer2015}}.

Acoustic analysis of speech can make use of ASR to count dysfluencies in spoken language that are often associated with neurodegenerative decline.
Pakhomov \etal made an early attempt to use ASR to extract many such prosodic features (pause-to-word ratio, normalized pause length, \etc) on picture-description task transcripts for participants with three variants of Frontotemporal Lobar Degeneration (FTLD)~\cite{pakhomovComputerizedAnalysisSpeech2010}.
A more recent pair of studies by T\'oth \etal explored using ASR for detection of MCI~\cite{tothAutomaticDetectionMild2015, tothSpeechRecognitionbasedSolution2018}.
However, in their work, only acoustic features were considered, and precise word transcripts were not required, mitigating the effect of the typically high WER for clinical speech samples.
Instead, the authors trained a new ASR model with a focus of detecting individual phonemes.
The features considered in this study were mostly prosodic (articulation rate, speech tempo, length of utterance, duration of silent and filled pauses, the number of silent and filled pauses).
The focus of the study was to compare the effects of manually annotating transcripts with the faster ASR method.
Since most ASR models cannot differentiate between filled pauses and meaningful voiced speech, their detection was a major focus of this work for automated MCI detection.
The ASR model was trained with annotated filled pause samples to learn to detect them in spontaneous speech.
The authors were able to show comparable results between the ASR and manual methods for MCI detection with the same feature set ($82\%$ accuracy for manual vs. $78\%$ for ASR)~\cite{tothAutomaticDetectionMild2015}.

While acoustic speech processing on its own has been less explored in detecting thought-disorder related mental illness, some researchers have found ways in which useful information can be derived solely from speech signals for this purpose.
One example is seen in work by Tahir \etal~\cite{tahirNonverbalSpeechAnalysis2016}.
In this study, patients with severe schizophrenia, receiving Cognitive Remediation Therapy (CRT), were differentiated from control participants with less severe schizophrenia (no CRT recommended) by \emph{non-verbal} speech analysis.
They note that nonverbal and conversational cues in speech often play a crucial role in communication, and that it is expected that individuals with schizophrenia would have a muted display of these features of speech.
Cues used as inputs to a classifier included interruptions, interjections, natural turns, response time, speaking rate, among others.
Preliminary results from this study with participants with severe schizophrenia ($n=8$) and less-severe forms of the disease ($n=7$) indicate that these nonverbal cues show approximately $90\%$ accuracy in classifying control participants from those with more severe forms of schizophrenia.
They also attempted to validate the computed features by examining their correlation with traditional subjective clinical assessments. 
Some of the computed objective nonverbal speech cue features had high correlation with subjective assessments; \eg ``poor rapport with interviewer'' has a strong correlation with longer participant response times.
The acoustics of bipolar disorder have also been studied, for example by Guidi \etal~\cite{guidiAutomaticAnalysisSpeech2015}.
In this study, the authors propose an automated method for estimating the contour of $F_0$ over time with a moving window approach as a proxy for mood changes.
In particular, they study local rising and falling events of the $F_0$ contour, including positive and negative slopes, amplitude, duration, and tilt to indicate different emotional states.
The features were first validated on a standard emotional speech database and then used to classify bipolar patients ($n=11$) and healthy control subjects ($n=18$).
They noted that intra-subject analysis showed good specificity in classifying bipolar subjects and healthy controls across all contour features, but that directions of most were not consistent across different subjects.
Due to limited data, they propose a study with a larger number of subjects including glottal, spectral, and energy features.

\subsubsection{Combination of acoustic and textual features}
Many dementia studies also use both acoustic and textual data  with promising results.
As an example, the previously mentioned work by Roark \etal (in Section~\ref{sec:nlp}) also made use of acoustic speech samples to aid in the detection of MCI from naturally-produced spoken language.
The researchers used manual and automated methods to estimate features related to the duration of speech during each utterance, including the quantity and duration of pause segments.
Some of the features that were computed include fundamental frequency, total phonation time, total pause time, pauses per sample, total locution time (both phonation and pauses), verbal rate, and several others~\cite{roarkSpokenLanguageDerived2011}.
They conclude that automated speech analysis produces very similar results to manually computing these metrics from the speech samples, demonstrating the potential of automated speech signal processing for detecting MCI.
Additionally, they found that a combination of linguistic complexity metrics and speech duration metrics lead to improved classification results.
The previously described work on PPA subtypes in~\cite{fraserAutomatedClassificationPrimary2012} was expanded by Fraser \etal in~\cite{fraserUsingTextAcoustic2013}.
Acoustic features were also extracted and added to the previous set of linguistic features to improve the classification results of PPA subtypes (PNFA and semantic dementia) and healthy control participants.
The added acoustic features included temporal prosodic features (\ie speech duration, pause duration,pause to word ratio, \etc), mean and variance of $F_0$ and first three formants ($F_1$, $F_2$, $F_3$), mean instantaneous power, mean and maximum first autocorrelation function, instantaneous power, and vocal quality features, \ie jitter and shimmer.
The authors tested the relative significance of all features using different feature reduction techniques and noted that more textual features were usually selected in each case.
However, the addition of acoustic features had the greatest positive impact when attempting to differentiate between healthy control participants and those with one of the PPA subtypes, but proved less useful in distinguishing the subtypes.
Their later study on AD~\cite{fraserLinguisticFeaturesIdentify2015} also used a similar hybrid approach with speech and language metrics to show good classification separating AD participants from healthy controls~\cite{fraserLinguisticFeaturesIdentify2015}.
The DementiaBank\footnote{{\label{footnote:DB}}{\url{https://dementia.talkbank.org/access}, accessed August 20, 2019}} corpus was used to collect the data for this analysis.
The study considered 370 distinctive features;
linguistic features included grammatical features (from part-of-speech tagging), syntactic complexity (\eg mean length of sentences, T-units, clauses, and maximum Yngve depth scoring for the parse tree, as described above), information content (specific and nonspecific word use), repetitiveness of meaningful words, and many more.
Acoustic features associated with pathological speech were also identified by computation of MFCCs, their derivatives, and their second derivatives.
To differentiate the clinical and control group, they considered mean, variance, skewness, and kurtosis of the MFCCs over time.
After performing factor analysis on these features, they showed that most of the variance between controls and those with AD could be explained by semantic impairment, acoustic abnormalities, syntactic impairment, and information impairment.

\subsubsection{Impact of ASR on textual features}
Several studies have also used ASR to generate transcripts of spoken language tasks for textual feature extraction for dementia detection.
However, unlike the phone-level ASR model built in~\cite{tothAutomaticDetectionMild2015} and~\cite{tothSpeechRecognitionbasedSolution2018}, this use case does require accurate word-level transcripts (\ie a low $\mathrm{WER}$).
Previous work has shown that ASR accuracy is reduced for both elderly patients and those with dementia~\cite{vipperla2008longitudinal, youngDifficultiesAutomaticSpeech2010, hakkani2010speech}.
To address this, Zhou \etal performed a study in which the DementiaBank\footnotemark[{\getrefnumber{footnote:DB}}] corpus was used to train an ASR model on domain data with elderly patients, both with and without AD~\cite{zhouSpeechRecognitionAlzheimer2016}.
They were able to show that an ASR model trained with a smaller in-domain dataset could improve $\mathrm{WER}$-based accuracy than one trained with a larger out-of-domain dataset.
Additionally, they were able to confirm that even with their model, diagnostic accuracy decreases with increasing $\mathrm{WER}$, as expected, but the correlation between the two is relatively weak when selecting certain features that are more robust to ASR errors (such as word frequency and word length related features)\footnote{The authors identify features that provide best diagnostic ability for gold-standard manual transcripts and transcripts with varying $\mathrm{WER}$ and ASR to identify these robust features, but they do not claim to understand why certain features seem more robust than others}.

Mirheidari \etal also used ASR with a combination of acoustic (temporal prosodic) and textual features (syntactical and semantic features) to diagnose and detect participants with neurodegenerative dementia (ND) and differentiate them from those with non-dementia related Functional Memory Disorder (FMD) with a conversational analysis dataset~\cite{mirheidariDiagnosingPeopleDementia2016, mirheidariAutomationDiagnosticConversation2017}.
With manual transcriptions, the classification accuracy was over $90\%$ in classifying the two groups, but it dropped to $79\%$ when ASR was used.
As expected, they found that the significance of the syntactic and semantic textual features is diminished when transcriptions contain ASR errors.
Sadeghian \etal attempted to improve the issue of transcription errors by training a custom ASR model using collected speech samples from participants with AD ($n=26$) and healthy controls ($n=46$) ~\cite{sadeghianSpeechProcessingApproach2017}.
This was done by limiting the potential lexicon to the collected speech in their dataset as well as cleaning the audio files to reduce the $\mathrm{WER}$.
Their study used a combination of acoustic features (temporal prosodic features and $F_0$ statistics) and textual features computed from both manual and ASR-generated transcripts (POS tags, syntactic complexity measures from~\cite{roarkSpokenLanguageDerived2011}, idea density, and LIWC features~\cite{tausczik2010psychological}).
In their work, the best classification results (over $90\%$) were seen when feature selection was performed using both the MMSE scores and computed acoustic and textual features, but using the computed features alone was nearly comparable and outperformed the MMSE scores on their own.
Weiner \etal~\cite{Weiner2017} instead compared the difference in the analysis of manual and ASR-derived transcripts for  a large range of acoustic (prosody and timing related) and textual features (lexical diversity with Brunet's index and Honor\'e's statistic) for comparing participants with dementia and healthy controls.
The off-the-shelf ASR model used in this work had a relatively high $\mathrm{WER}$, but they were interestingly able to show that the $\mathrm{WER}$ itself was a reliable feature for classifying the different types of subjects.
Additionally, many of the features they selected showed robustness to transcription quality, possibly even taking advantage of the poor ASR performance to identify participants with dementia.

\subsection{Advantages \& Disadvantages}
It is intuitive that fine-grained and discrete measures of ``what is said'' (language, in terms of lexical diversity, lexical density, semantic coherence, language complexity, \etc) may more directly capture early cognitive-linguistic changes in illness and disease than measures of ``how it is said'' (analysis of speech acoustics).
However, emerging data shows that acoustic analysis offers converging and complementary information to several of the textual features discussed in Section~\ref{sec:nlp}.
Most interestingly, changes in the outward flow of speech may precede measurable language-based changes~\cite{tahirNonverbalSpeechAnalysis2016, fraserUsingTextAcoustic2013}.

A particular advantage of evaluating speech acoustics is that ASR or transcription is not necessarily a required step.
Automated acoustic metrics can be extracted from non-labeled speech samples~\cite{konigAutomaticSpeechAnalysis2015,tahirNonverbalSpeechAnalysis2016, lopez-de-ipinaAutomaticDiagnosisAlzheimer2015, konigFullyAutomaticSpeechBased2018}.
Further, some of the metrics provide complementary and interpretable value that cannot be gleaned from transcripts (rate, pause metrics, speech prosody).
These directly correspond with subjectively described clinical characteristics (\eg pressed speech, halting speech, flat affect etc).
A disadvantage is that not all acoustic metrics offer that level of transparency. This is a running theme in clinical speech analysis.
Many of these features are not currently used in clinical diagnosis despite their powerful predictive power because they are difficult to directly interpret (\eg MFCCs); this means that clinicians can see the output of a complicated model but not understand why the model came to that decision or if it is considering clinically-relevant dimensions.
For this reason, some effort has been undertaken to map the information contained in high-dimensional data to be easily visualized and interpreted by clinicians, but this remains a significant challenge~\cite{jiaoInterpretablePhonologicalFeatures2017, tuInterpretableObjectiveAssessment2017}.
\section{Concluding Remarks and Future Work}\label{sec:discussion}
An analysis of the existing literature reveals a set of future research directions to help advance the state of the art in this area. 
Here, we provide an overview of these directions and highlight some important open questions in this space.

\subsection{Characterizing Inter and Intra-Speaker Variability in Healthy Populations}
There is a great deal of variability to be expected in speech and language data.
Extensive work on the language variables influencing inter- and intra-speaker variation suggest that any level of language (\ie phonological, phonetics, semantics, syntax, morphology) is subject to both conscious/explicit and completely unconscious/subtle variation within a speaker.
These conscious and unconscious sources of variability are conditioned by pragmatics, style-shifting, or register shifting~\cite{coupland2007style, schilling2013investigating}.
Similarly, speech acoustics are impacted by speaker identity, context, background noise, spoken language, \etc~\cite{benzeghibaAutomaticSpeechRecognition2007}.
These individual sources of variability have yet to be fully characterized quantitatively for the features that we described in this paper.
A more complete understanding of this variability in healthy populations helps to interpret changes observed in clinical populations.
For example, this knowledge can help understand how typical or atypical is a particular semantic coherence score (\eg in what percentile does the semantic coherence score fall?).
Furthermore, this understanding can inform stratified sampling schemes that allow experimenters to match healthy and clinical cohorts on relevant aspects of speech/language production. This is critical for clinical trial design.

\subsection{Joint Optimization of Speech Elicitation and Speech \& Language Analytics}
Algorithms published in the literature typically make use of previously-collected speech and language samples.
These samples are often collected for other reasons and are only used by algorithm designers because they are available.
A related challenge also arises in comparing the results of various analyses of speech and language analytics for cognitive assessment when the datasets used in each study are vastly different.
This is evident in the large range of speech elicitation tasks used in many of the studies mentioned in this review.
The lack of standardization in data collection for cognitive assessment therefore contributes to the problem of limited available data for any particular task.

As a result, published results are potentially biased because these data sets are small and collected on a limited set of elicitation tasks.
Deeper collaborations between speech neuroscientists, neuropsychologists, and speech technologists are required to push the state-of-the-art forward.
There is an extensive literature on how to efficiently and reliably elicit speech to exert pressure on the underlying cognitive-linguistic processing~\cite{muellerLatentStructureTest2018}.
The algorithms for extracting clinically-relevant information from speech and audio have been developed independently from this work.
We posit that joint exploration of the elicitation-analytics space has the potential to result in improved sensitivity in detecting cognitive-linguistic change. 
This analysis will reveal which elicitation tasks and analytics are maximally sensitive for a given problem. 

The rate-limiting factor in being able to answer these questions is the lack of publicly-available data on a large scale. Unlike speech recognition research, data for clinical speech research is much more difficult to collect and share because it's scarcer, requires experts to label, and there are privacy issues. Until these issues are resolved, it will be difficult to overcome some of the scientific challenges we highlight here.

\subsection{Robustness to Noisy Data}
The sensitivity of the features we describe herein and the follow-on models they drive are not well understood under noisy conditions. 
Our definition of \emph{noisy} is rather loose here.
For example, noise may arise from imperfect transcripts provided by an ASR engine, background noise that may corrupt the acoustics, or feature distribution mismatch between training and test data in supervised settings.
Unimportant nuisance parameters for clinical applications (\eg idiosyncratic features related to different speakers) are especially problematic in acoustic analysis~\cite{benzeghibaAutomaticSpeechRecognition2007}.
A better characterization of the sensitivity of these nuisance features can inform the development of new representations that are robust to various sources of noise.
These models can improve the algorithms' ability to generalize and can help understand the fundamental limits of speech as a diagnostic.

\subsection{Data-Driven and Interpretable Features}
Many of the features described herein are readily interpretable and, given the existing literature, it is reasonable to posit that they have clinical utility.
However, if clinical speech data becomes available on a large scale, we expect that data-driven artificially intelligence (AI) systems will replace some of the domain-expert features described herein.
For example, it is reasonable to expect that features that are optimized for a specific application (\eg diagnosing schizophrenia) would outperform the general-purpose features described here.
This improved performance likely comes at the expense of reduced feature interpretability. 
An area ripe for further exploration in clinical speech analytics, and clinical analytics in general, is the development of AI models that provide interpretable outputs when interrogated, such as in \cite{tuInterpretableObjectiveAssessment2017}. This area has received some attention recently and will continue to become more important as AI systems are deployed in healthcare.

\bibliographystyle{IEEEtran}
\bibliography{references/jstsp_refs}

\IEEEpeerreviewmaketitle

\begin{IEEEbiography}[{\includegraphics[width=1in,height=1.25in,clip,keepaspectratio]{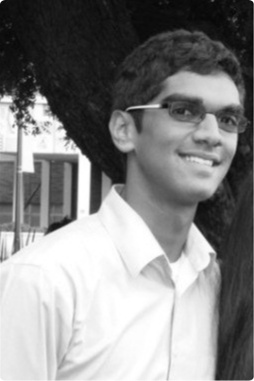}}]{Rohit Voleti}
is a Ph.D. student in the School of Electrical, Computer, \& Energy Engineering at Arizona State University (ASU) in Tempe, AZ, USA. Prior to his time at ASU, he obtained a B.S. and M.S. in Electrical Engineering from the University of California, Los Angeles (UCLA), and worked as a systems engineer in the medical device industry in Southern California.
\end{IEEEbiography}

\begin{IEEEbiography}[{\includegraphics[width=1in,height=1.25in,clip,keepaspectratio]{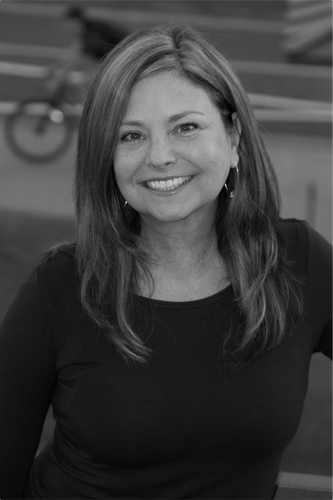}}]{Julie M. Liss}
is a Professor of Speech \& Hearing Science and Associate Dean of the College of Health Solutions at Arizona State University (ASU) in Tempe, AZ, USA. Her research explores the ways in which speech and language change in the context of neurological damage or disease.
\end{IEEEbiography}

\begin{IEEEbiography}[{\includegraphics[width=1in,height=1.25in,clip,keepaspectratio]{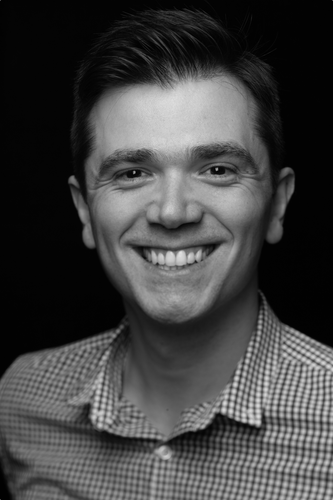}}]{Visar Berisha}
is an Associate Professor in the College of Health Solutions and Fulton Entrepreneurial Professor in the School of Electrical Computer \& Energy Engineering at Arizona State University (ASU) in Tempe, AZ, USA. His research interests include computational models of speech
production and perception, clinical speech analytics, and statistical signal processing.

\end{IEEEbiography}

\vfill

\end{document}